\documentclass[final]{cvpr}

\usepackage{times}
\usepackage{epsfig}
\usepackage{graphicx}
\usepackage{amsmath}
\usepackage{amssymb}
\usepackage{subfig}

\usepackage{tabularx, booktabs}
\usepackage{amsmath}
\usepackage{amssymb}
\usepackage{pifont}
\usepackage{threeparttable}
\usepackage{color}
\usepackage{enumitem}
\usepackage{microtype}
\usepackage{etoc}

\definecolor{mygreen}{RGB}{20, 190, 10}
\definecolor{myorange}{RGB}{250, 86, 36}
\definecolor{myblue}{RGB}{100, 100, 255}

\newcommand{\ignore}[1]{}
\newcommand{\titlecap}[2]{\textbf{#1} #2}
\newcommand{\new}[1]{\textcolor{black}{#1}}

\usepackage[pagebackref=true,breaklinks=true,colorlinks,bookmarks=false]{hyperref}

\def\cvprPaperID{576} 
\def\confYear{CVPR 2021}

\newcommand{\mypara}[1]{\vspace{1mm}\noindent \textbf{#1}}

\definecolor{darkred}{rgb}{0.55, 0.0, 0.0}

\definecolor{dimgray}{rgb}{0.41, 0.41, 0.41}

\newcommand{\para}[1]{\noindent\textbf{#1}}

\newcommand{\bi}{\begin{itemize}}
\newcommand{\ei}{\end{itemize}}

\begin{document}

\etocdepthtag.toc{mtchapter}
\etocsettagdepth{mtchapter}{subsection}
\etocsettagdepth{mtappendix}{none}

\title{Compositionally Generalizable 3D Structure Prediction}

\author{Songfang Han\\
UC San Diego\\
\and
Jiayuan Gu\\
UC San Diego\\
\and
Kaichun Mo\\
Stanford University\\
\and
Li Yi\\
Google Research\\
\and
Siyu Hu\\
USTC\\
\and
Xuejin Chen\\
USTC\\
\and
Hao Su\\
UC San Diego\\
}

\twocolumn[{%
\renewcommand\twocolumn[1][]{#1}%
\maketitle
\begin{center}
    \centering
    \vspace{-8mm}
    \captionsetup{type=figure}
    \includegraphics[width=\textwidth]{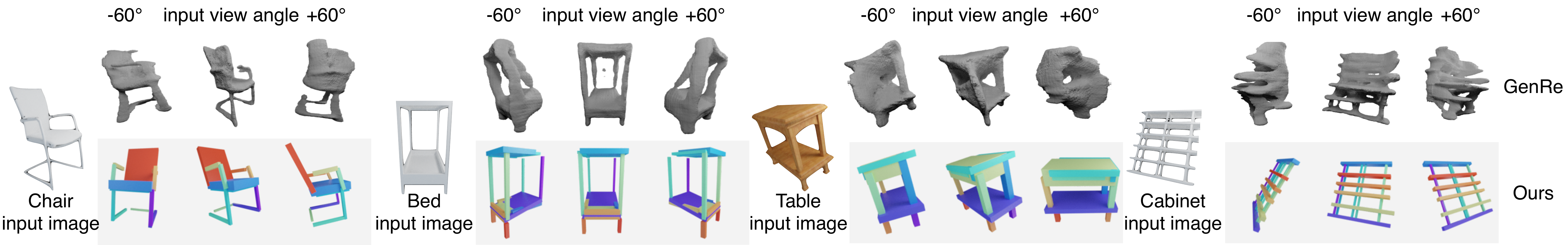}
    \captionof{figure}{3D structure predictions using models trained on \emph{chair} shapes, by GenRe~\cite{zhang2018learning} and our approach (visualized at three views).  Our method can generalize to novel unseen categories: bed, table, and cabinet. We achieve superior reconstruction results than GenRe. }
    \label{fig:teaser}
\end{center}%
}]


\begin{abstract}
Single-image 3D shape reconstruction is an important and long-standing problem in computer vision.
A plethora of existing works is constantly pushing the state-of-the-art performance in the deep learning era.
However, there remains a much more difficult and under-explored issue on how to generalize the learned skills over
unseen object categories that have very different shape geometry distributions.
In this paper, we bring in the concept of compositional generalizability and propose a novel framework that could better generalize to these unseen categories. We factorize the 3D shape reconstruction problem into proper sub-problems, each of which is tackled by a carefully designed neural sub-module with generalizability concerns.
The intuition behind our formulation is that object parts (\eg slates and cylindrical parts), their relationships (\eg adjacency and translation symmetry) and shape substructures (\eg T-junctions and a symmetric group of parts) are mostly shared across object categories, even though object geometries may look very different (\eg chairs and cabinets).
Experiments on PartNet show that we achieve superior performance than state of the art. This validates our problem factorization and network designs.
\end{abstract}

\section{Introduction}
Single-image 3D reconstruction is a critical and long-standing problem in computer vision. 
Existing works in the literature of 3D deep learning~\cite{Choy:2016,Fan:2017,Pixel2Mesh,Chen:2019,Mescheder:2019,Park:2019} mostly 
focus on designing better network architectures for decoding 3D geometry from 2D image inputs, assuming that the training and test data come from the same or similar categories.

We are interested in a much harder setting, where the training and testing data may be sampled from very different object categories. 
An example would be to reconstruct beds after seeing training data of chairs. 
While such cross-category generalizability is under-explored in the 3D deep learning literature, it is a fundamental capability of humans --- as we explore 2D product images online, we constantly imagine 3D geometry of objects from unknown categories yet seldom running into difficulties.

Very few works investigated this cross-category generalization setup~\cite{zhang2018learning,wanggsir}.
Inspired by David Marr's seminal idea on 2.5D representations, GenRe~\cite{zhang2018learning} and its follow-up work~\cite{wanggsir} hypothesize that depth map in image space leads to a better cross-category generalizability than predicting 3D shapes directly. 
While it does generalize better than previous works, its basis, generalizable depth prediction across categories, is indeed an unsolved and difficult problem.
\new{Besides, merely using depth representation fails to leverage the underlying structure of objects, which is critical and natural to human understanding of 3D shapes.
For example, humans can reconstruct a chair as a composite of several similar chair legs.}
Fig.~\ref{fig:teaser} presents exemplar comparisons between GenRe~\cite{zhang2018learning} and our method on the cross-category 3D reconstruction, where we clearly observe that GenRe suffers from imperfect depth estimation, while our method can recover the object structures more faithfully.

In this work, we explore a very distinctive path toward cross-category generalizability based on the following observation: while training and test objects from different categories might follow very different whole-object level shape distribution, they often share similar distribution in terms of ``substructures''. 
For example, among all man-made objects, slates and cylindrical parts are quite common. And among all possible configurations of parts, parallel and T-junction relationships are more frequently observed. 
Taking the Bayesian's view, the joint probability $\Pr(p_1, p_2, \ldots, p_n)$ of all substructures ${p_i}$ could be very different across training and test sets. However, if we factorize the joint probability as products of low-order potentials $\phi(p_i)$ and $\phi(p_i, p_j)$, these potentials are often much better aligned. 
This motivates us to factorize the problem of shape reconstruction into subproblems of substructure prediction, which has the potential of being more generalizable across categories. We refer this factorization-enabled generalizability as \emph{compositional generalizability}~\cite{mu2020refactoring,loula2018rearranging}.




\new{To this end, we interpret a 3D shape as a composite of parts.
Specifically, we use \textbf{oriented bounding boxes} to represent parts and parameterize a 3D shape by \emph{connectivity}, \emph{orientation}, and \emph{size} of part bounding boxes. Instead of focusing on fine geometry, this representation encodes abstract structure, which facilitates many applications that require structure-level reasoning ~\cite{tulsiani2017learning,mo2019structurenet,mo2020structedit}.}

\new{To predict such structured shapes, we build a system composed of independent modular networks.
We first extract part segmentation from an input image. The part mask provides an explicit attention map. The network is encouraged to learn local structures, by keeping only the part masks of interest in the input. Given part masks, we further factorize shape reconstruction into a series of sub-problems, that are carefully designed and demonstrated with strong transferability. In particular, we predict the part orientation for each part mask and then regress the size in a group-wise fashion. After obtaining a set of part bounding boxes, we finally learn to assemble these part bounding boxes into a 3D shape.
To tackle non-trivial challenges for independent modules, we also devise several novel techniques.}


\new{For the part orientation prediction module,  we observe that the ground truth rotation annotation suffers from unavoidable ambiguity, which is caused by the ambiguity of principal axes and by part geometric symmetry. To address this issue, our approach predicts principal axes of oriented bounding boxes in an orderless fashion. Additionally, a mixture of experts is applied to predict multiple rotations and the confidence to select each. The association of multiple experts further improves the performance. From our experiments, the proposed method accelerates training and yields a much smaller angle error.}

\new{
To address the challenge of ill-posed size prediction from a single image, we estimate the part size in a group-based manner. We first group translational symmetric parts and then aggregate part features in each group to predict a unified size. Such group-level information sharing allows better reasoning and more accurate predictions, especially when perspective distortion exists.
}

\new{For the position prediction, we infer the relative position between two parts instead of the absolution position. Compared to the absolute translation of each part used in~\cite{li2020learning}, this relative relationship shows stronger cross-category transferability.}

In summary, the key contributions in this paper are listed as follows. 
First, 
we carefully design a modularized pipeline to solve each of these sub-problems in a transferrable manner. 
\new{Second, we introduce a novel orientation prediction module to address the intrinsic symmetry ambiguity issue with a mixture of experts.}
\new{Third, a group-based size prediction module is introduced to mitigate the size ambiguity caused by image projection. } 
Fourth, we demonstrate significant improvement over prior arts and provide extensive analysis to justify our designs. 


\begin{figure*}[!ht]
    \centering
    \includegraphics[width=\textwidth]{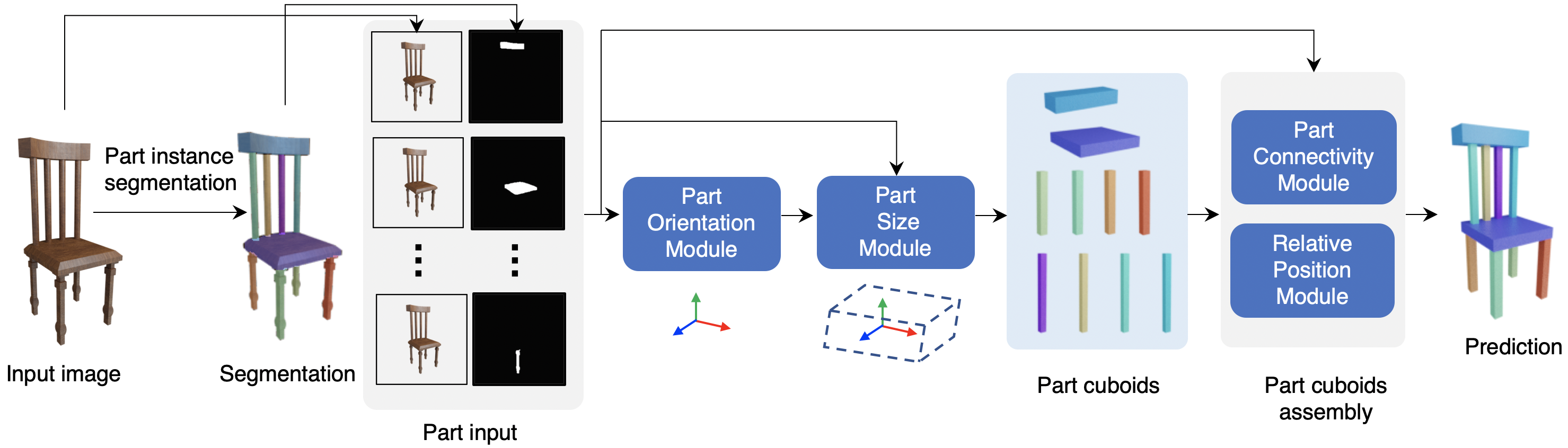}
    \caption{\textbf{Our single-view 3D structure prediction pipeline.} To extract the 3D structure from a single input image, we first apply MaskRCNN to extract the part instance masks. Then we predict each part cuboid attributes including orientation and size. Finally, we predict the pairwise connectivity and assemble the part cuboid together through relative position. Note that we heavily use part masks as module inputs to induce a focus on local regions. }
    \label{fig:overview}
\end{figure*}
\section{Related Work}
We review previous works on reconstructing 3D shape from single images and part-based shape modeling, along with discussions and comparisons to our approach.

\mypara{Single-image 3D Shape Reconstruction.}
Estimating 3D geometry from 2D single images is a long-standing problem in computer vision.
Due to its ill-posed nature, learning-based methods~\cite{Choy:2016,Tatarchenko:2017,Wang:2018:ocnn,Richter:2018,Fan:2017,Lin:2018,Insafutdinov:2018,Pixel2Mesh,Pixel2Mesh++,Groueix:2018,Chen:2019,Mescheder:2019,Park:2019,Saito:2019,Xu:2019} are proven effective recently since they can learn shape priors from training data.
However, most of these works assume the alignment of the training and testing data distributions, \eg either from the same object category or among multiple similar categories.
Even though some works, such as ~\cite{Fan:2017}, show that their methods, if trained in a category-agnostic manner over a large set of object categories, can reconstruct 3D shapes from novel categories, there are no explicit designs to guarantee such desired properties.
Tatarchenko~et~al.~\cite{Tatarchenko:2017} demonstrated that state-of-the-art single-image 3D shape reconstruction methods tend to memorize the training shapes and perform retrieval at the testing time.
In our work, we design explicit network modules to discover part relationships and shape sub-structures that are shared across different object categories to gain better reconstruction generalization across category gaps.

It is an under-explored problem to extrapolate the reconstruction capabilities over out-of-the-distribution novel categories.
GenRe~\cite{zhang2018learning} proposes to factorize the problem into a sequence of sub-tasks -- depth estimation, depth spherical map inpainting, voxel-based 3D refinement, and observes better generalization capability to unseen testing object categories.
GSIR~\cite{wanggsir} further extends the GenRe approach by bringing in the part structure information.
Different from these two works that leverage depth as intermediate steps for generability,
our approach learns to discover part relationships and shape sub-structures that are commonly shared among man-made object categories.
Wallace~et~al.~\cite{wallace2019few} and Michalkiewicz~et~al.~\cite{michalkiewicz2020few} propose few-shot learning frameworks to quickly adapt the learned shape priors to novel categories.
Our method uses a zero-shot setting that no data from the testing categories is needed at all.

\mypara{Part-based Shape Structure Modeling.}
Most 3D shapes are composed of sub-structures and parts.
While most previous works mentioned above reconstruct 3D shape geometry as a whole entity, understanding shape structures and parts is a crucial perception task for robotic manipulation~\cite{katz2008manipulating,martin2019rbo,xiang2020sapien}, shape generation~\cite{Li:2017,Dubrovina:2019,Schor:2019,Wu:2019:pqnet,Gao:2019,mo2019structurenet,Li:2020,jones2020shapeassembly,tian2019learning} and editing~\cite{kraevoy2008non,xu2009joint,mo2020structedit,sung2020deformsyncnet}.

There are many recent works that learn to abstract shape parts using primitive shapes, such as cuboids~\cite{tulsiani2017learning,zou20173d,sun2019learning}, superquadratics~\cite{paschalidou2019superquadrics,paschalidou2020learning}, mixtures of Gaussians~\cite{genova2020local,genova2019learning}, convex hulls~\cite{deng2020cvxnet}, etc.
However, these works do not explicitly consider part relationships, such as the part adjacency and symmetry constraints.
In this paper, we explicitly discover and model such part relationships not only for better reconstruction results, but also for generalizing to novel categories that share similar part structures.

Previous works studied various ways for modeling part relationships and constraints.
Early works~\cite{Chaudhuri:2011,Kalogerakis:2012,Jaiswal:2016} studied pairwise part relationships using probabilistic graphical models.
More recently, GRASS~\cite{Li:2017}, Im2Struct~\cite{niu2018im2struct} and StructureNet~\cite{mo2019structurenet} organize parts in tree structures and also model part adjacency and symmetry constraints among sibling parts.
SAGNet~\cite{Wu:2019:sagnet} and SDM-Net~\cite{Gao:2019} also design explicit network modules to learn pairwise part relationships for shape generation.
Recent papers~\cite{li2020Learning3DPart,HuangZhan2020PartAssembly} learns part relationships for part assembly via graph neural networks.
Though working well within a collection of shapes from the same object category or very similar categories, most of these works failed to generalize to novel categories, which is the central goal of our work.

\section{Problem Statement}
\label{sec:problem}
Given an image $I$ of an object $O$, our goal is to predict the 3D structure of $O$, which is represented as an assembly of geometric primitives $\{p_i\}$. In our case, we use the oriented bounding box as primitive following StructureNet~\cite{mo2019structurenet}, \ie, $p_i=[c_x, c_y, c_z, s_x, s_y, s_z, q]$, where $(c_x, c_y, c_z)$ denotes the center location, $(s_x, s_y, s_z)$ represents the length of each edge, and $q \in \mathbb{R}^4$ corresponds to the quaternion representation of the rotation. Especially, we focus on the relative position between a pair of parts $(c_x^i - c_x^j, c_y^i - c_y^j, c_z^i - c_z^j)$.
In our case, an object-level shift of center location is allowed, as long as the relative positions between pairs keep. 

The learning of all the modules in our work are supervised. During training, the groundtruth part instance segmentation masks and the ground-truth parameters of the oriented bounding boxes are provided. In practice, the images are rendered from the CAD models in ShapeNet~\cite{chang2015shapenet}, and the ground-truth part decomposition and part parameters are generated according to 3D labels in PartNet~\cite{mo2019partnet}. The oriented bounding box of a part are computed using Pinciple Component Analysis (PCA) according to its part geometry. 

\section{Method}
Compared to previous methods, our proposed algorithm is trained only on the chair category, yet can generalize to unseen categories, such as bed, storage furniture, and table. The key observation is that shapes across categories share similar parts and local part relationships, supporting compositional generalizability.

The overall framework is summarized in Fig.~\ref{fig:overview}.
Given an image, we first learn to predict part masks using MaskRCNN~\cite{he2017mask}, a well-established object instance segmentation approach (Sec~\ref{sec:method:part_seg}). In Sec~\ref{sec:method:part_ori_pred} and Sec~\ref{sec:method:part_scale_pred}, we introduce how to predict the direction and size of the oriented bounding box for each part. Finally, in Sec~\ref{sec:method:part_assembly}, we introduce how to assemble the predicted boxes into a complete shape.

\subsection{Part Instance Segmentation} 
\label{sec:method:part_seg}
The first step of our pipeline is to obtain the mask of each part in the input image. The goal of this step is exactly the same as the classical instance-level object segmentation problem. Due to the popularity of research into instance-level segmentation, we do not treat this step as the focus of our work. Instead, we borrow existing methods to obtain the part masks. Particularly,  we choose the well-established MaskRCNN~\cite{he2017mask} and train it using the finest part masks rendered from PartNet. 
Please refer to Sec~\ref{sec:supp_part_instance_seg} in the supplementary material for details.

  
\subsection{Part Orientation Prediction}
\label{sec:method:part_ori_pred}

\begin{figure}[t]
    \centering
    \includegraphics[width=\linewidth]{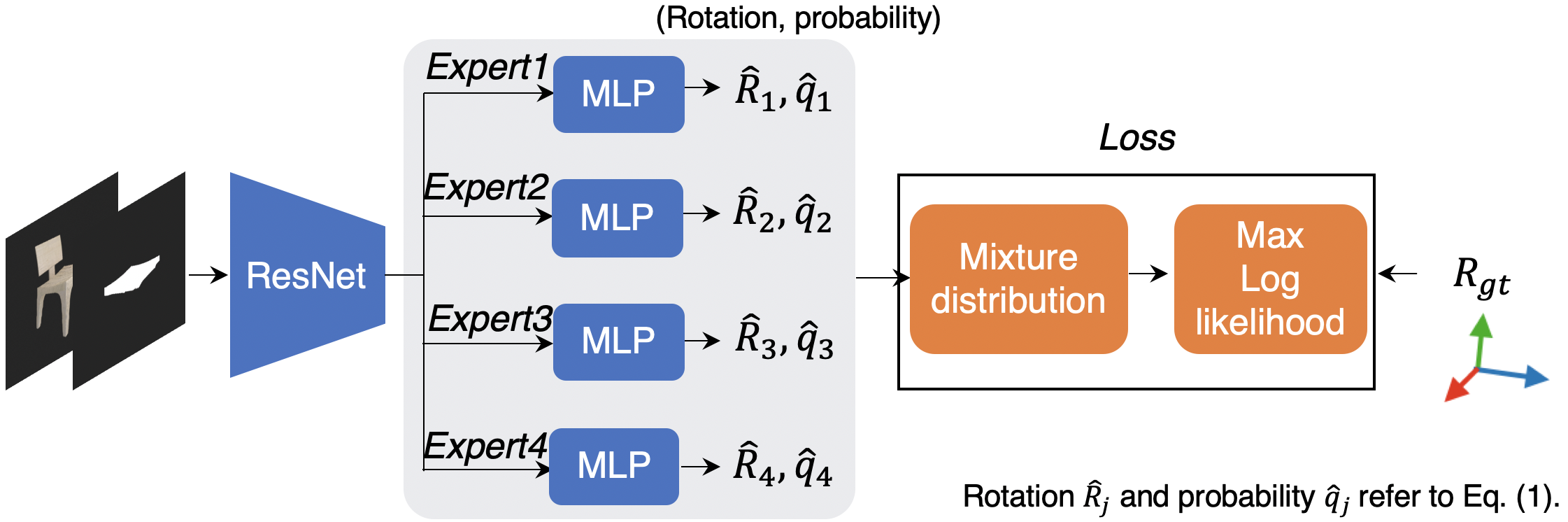}
    \caption{\textbf{Part orientation prediction module.} 
    Given an RGB image and a part mask, we extract the global feature, from which the rotation is predicted. Specially, we predict 4 possible rotations as well as their probabilities with the Mixture of Experts.
    \new{At the training stage, we maximize the log-likelihood under the rotation distribution. During inference, the rotation with the highest probability is selected.}
    }
    \label{fig:part_ori_module}
\end{figure}

\new{Fig~\ref{fig:part_ori_module} illustrates our part orientation prediction module, which consumes an RGB image and a binary part mask to estimate the part orientation.
Specifically, a ResNet is employed to extract a global feature from the concatenation of an image and a binary mask. Then this latent feature is fed to a multi-layer perceptron which regresses the orientation.}

\mypara{Ambiguity of GT labels.}
\new{Since it is difficult to define a canonical pose for various parts, most existing part-level datasets, like PartNet, do not provide oriented bounding box annotations.
They are usually computed by the PCA given GT part segmentation.
Thus, such generated labels are inevitably imperfect due to ambiguity of principal component axes and underlying part symmetry.
For example, there exist 24 equivalent rotations for an equant part (cube).
In addition, the PCA is also sensitive to noise in GT segmentation labels.}

\mypara{Orderless Principal Axes Prediction.}
Naively supervising the network with ambiguous and inconsistent ground truth would hinder the training seriously as illustrated in Table~\ref{tab:ablate_rot}.
A straightforward solution is to introduce symmetry-aware losses~\cite{wang2019normalized} to deal with ambiguity caused by symmetry.
However, due to lack of GT symmetry annotations, heuristics are needed to distinguish different cases (equant, prolate, oblate, bladed), which is shown to be suboptimal in Sec~\ref{sec:exp:ablation}.

To this end, we propose a novel approach to get rid of the orientation ambiguity by estimating the set of principal axes without differentiating the order and the signs.
The output of our module is a rotation matrix\footnote{The network outputs a quaternion, which is converted to a rotation matrix. Other representations~\cite{zhou2019continuity}, are also tried and show similar results.}, each column of which is treated as a principal axis.
We consider two rotation matrices to be equivalent if the corresponding set of principal axes are the same without differentiating the signs.
We supervise the network predictions through a Min-of-N MSE loss with all equivalent rotation matrices of the GT one which form a GT set.
The loss is designed as the minimal mean squared error between the prediction and each rotation matrix in the GT set, denoted by $D(\hat{R}, \{R\})$.
Here $\hat{R}$ is the prediction, and $\{R\}$ is the GT set.
In Sec~\ref{sec:method:part_scale_pred}, we will discuss how to predict part sizes given a set of principal axes.

\mypara{Mixture of Experts for Rotation Symmetry and Multi-Covering of $\mathbb{SO}(3)$.}
However, the orderless principal axes prediction is not enough to solve the problem nicely in practice.
Also, according to \cite{xiang2020revisiting}, it is impossible to find a continuous map from Euclidean space to $\mathbb{SO}(3)$ when symmetry exists.
To this end, we introduce \emph{Mixture of Experts} (MoE)~\cite{jacobs1991adaptive}.
Concretely, given the features extracted by the backbone, multiple branches (a.k.a. ``experts'') are applied to predict multiple 
rotations $\{\hat{R}_j\}$
as well as probabilities $\{\hat{q}_j\}$ to select certain branches.
The loss is then updated as:
\[L_1=\min_{j}D(\hat{R}_j, \{R\}).\]

Besides, we also maximize the log-likelihood of the predictions under a mixture of Laplacian distributions, which is formulated as:
\begin{equation}
L_2=\log \sum_{j} \hat{q}_j \frac{1}{2b} e^{-\frac{D(\hat{R}_j, \{R\})}{b}}.
\end{equation}

The overall training loss is $L_1+\lambda L_2$.
During inference, we select the branch with the maximum probability $\hat{q}_j$ to predict $\hat{R}_j$.  
We summarize the innovations of our design here and include details in Sec~\ref{sec:supp_edge_direction} of the supplementary material. We provide qualitative results of rotation prediction in supplementary material Fig.~\ref{fig:rot_module}.


\subsection{Part Size Prediction} 
\label{sec:method:part_scale_pred}
\new{
Since a set of principal axes (order-invariant) is estimated by our part orientation prediction module, we predict the part size in an orderless fashion accordingly.
Concretely, the edge length of the bounding box along each principal axis is predicted.
In addition, we also propose group-based size prediction to mitigate the size ambiguity of a single image caused by perspective projection.
}

\begin{figure}[t]
    \centering
    \includegraphics[width=0.95\linewidth]{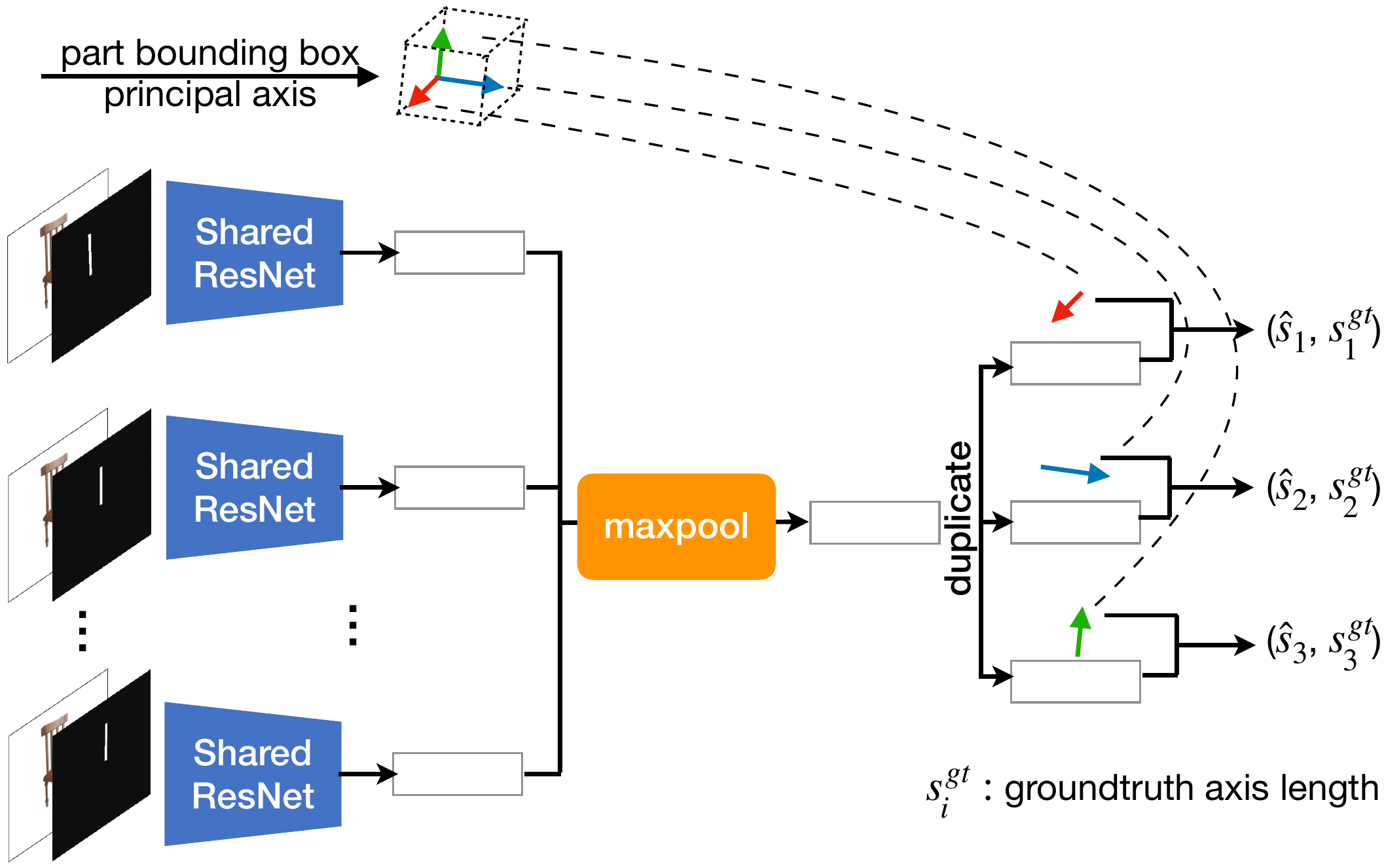}
    \caption{\textbf{Part size prediction module.}
    The module is designed to predict the size (edge length of bounding box) of parts in each translational symmetric group. We first extract a per-part feature with a shared ResNet and then aggregate part features through the max-pooling operation. A shared size is predicted given the aggregated group feature.}
    \label{fig:edge_length_pred}
\end{figure}

\mypara{Group-based Size Prediction.}
In this work, we form part groups by considering translational symmetry.
Specifically, all the parts are grouped into several clusters, so that parts in the same cluster can be translated to each other. The clustering scheme will allow us to predict a unified size for all the member parts in each cluster. Predicting part sizes in clusters brings at least three advantages: (1) Compared with predicting part sizes individually, we have more evidences from the input. Particularly, it significantly helps reduce the ambiguity of object sizes under perspective projection; (2) Some parts may not be fully visible due to occlusions, and predicting their sizes individually is very difficult; (3) Having a unified size for translational symmetric parts is visually much more pleasing for humans.
\new{The benefit of group-based prediction over unary size one is shown in Table~\ref{tab:ablate_group_size}.}

\new{
The architecture of our group-based size prediction is illustrated in Fig.~\ref{fig:edge_length_pred}. 
We first use a shared ResNet
to extract features for each part in a cluster independently.
The input of each part is a concatenation of an image and a part mask.
The mask induces the network to focus on the masked region. We then use a max-pooling layer to aggregate features of individual parts. Finally, the aggregated feature is duplicated three times and coupled with each principal axis predicted by the orientation prediction module (Sec~\ref{sec:method:part_ori_pred}).
To output the length along each axis, which should be invariant to the order of axes, we follow the segmentation network in PointNet~\cite{qi2017pointnet}.
During training, each principal axis is just acquired from the ground truth.
The module is supervised by the mean squared error.
 We summarize our design here and include details in Sec~\ref{sec:supp_edge_length} of the supplementary material. 
We provide qualitative results in supplementary material Fig.~\ref{fig:size_module}.
}


\mypara{Translational Symmetry Classifier.}
\new{
We train a translational symmetry classifier to help form clusters. The classifier takes an image and a pair of part masks, to predict whether these two parts highlighted by the input masks have translational symmetry. At test time, we build clusters of parts so that the symmetry score of any part pair in the same cluster is above a predefined threshold.
}

\subsection{Part Assembly} 
\label{sec:method:part_assembly}
Having computed part mask, orientation, and size of individual parts, this module learns to assemble them into a whole shape. We predict the connectivity relationship and assemble adjacent parts by predicting the relative position between them.

Our relative position based assembly pipeline is significantly different from most work in the 3D deep learning literature.  Most previous work estimates absolute location in the camera space, the world space, or a canonical object space. However, absolute locations are sensitive to the shape scale as well as the translation along the optical axis, significantly hurting the performance even for the simpler within-category prediction setup. While shape-level canonicalization is often conducted to mitigate the issue in literature, for unknown object categories, any prediction involving the undefined ``canonical object space'' is ill-posed.

\mypara{Connectivity-based Part Tree.} 
We train a connectivity classifier to assess whether a pair of parts touch each other in its original 3D shape.  Based upon the prediction of connectivity scores between all part pairs, we heuristically sample a part tree from the connected parts. Specifically, we choose the largest part (by predicted volume) as the root node. Next, we iteratively add more parts by selecting the largest remaining part in the 1-hop neighborhood of the current tree. 

\mypara{Joint-based Relative Position.}
We assemble the adjacent parts by predicting the relative position. The input includes the reference image and the information about each individual part in the pair. 

\begin{figure}[t]
    \centering
    \includegraphics[width=\linewidth]{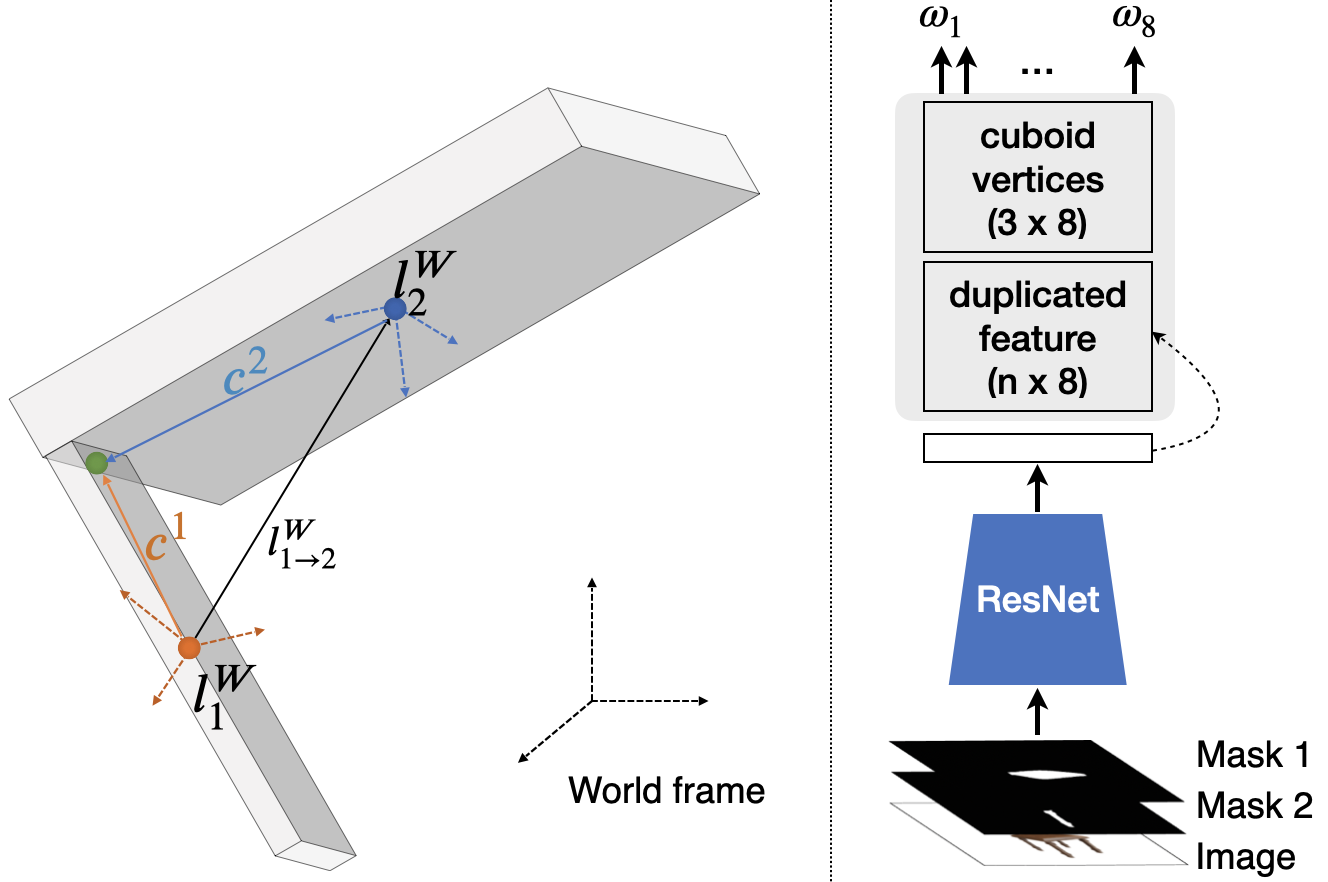}
    \caption{\textbf{Relative position prediction.} Left: Parameterize relative position between part centers (orange and blue spheres) by contact points (green sphere). Right: Network architecture illustration.}
    \label{fig:pos_pred}
\end{figure}

Instead of directly predicting relative center positions, we use a contact point based approach, which incorporates stronger relative position prior. Specifically, the contact point between touching part pairs must lie within the oriented bounding box of each part.
In the following, we formulate how to infer the relative center position from contact point predictions~(read Fig.~\ref{fig:pos_pred} for better understanding): 

\begin{itemize}[leftmargin=+0.1in]
    \item \para{Contact Point.} We use superscript to denote frame and subscript to denote part indices. Given two touching parts $p_1$ and $p_2$, we denote the contact point in a part frame sitting at the center of $p_1$ by $c^1$, and in a different part frame sitting at the center of $p_2$ by $c^2$.\footnote{The orientation of the two frames can be arbitrarily chosen.} We denote the part centers in the world frame as $l^W_1$ and $l^W_2$. Because the contact point is the same, we have the following constraint:
    \begin{equation}
    	l^W_1 + c^1 = l^W_2 + c^2
    \end{equation}
    The relative center location $l^W_{1\rightarrow_2}$ can be inferred as:
    \begin{equation}
    	l^W_{1\rightarrow_2} = l^W_2 - l^W_1 = c^1 - c^2.
    \end{equation}
    Next we discuss how to estimate the contact point $c^i$ in part frame.
    \item \para{Contact Point Estimation.} As the contact should be lying on the part surface or inside, we denote the contact points $c^i$ as a linear interpolation of the oriented bounding box vertices $v_{i, \cdot}$: $c^i = \sum\limits_{j=1}^8 \omega_{i,j} \cdot v_{i,j}$, where $\sum\limits_{j=1}^8 \omega_{i,j} = 1$ and $\omega_{i,j} \geq 0$. 
    We use a neural network to predict $\omega_{i,j}$'s. This network uses a softmax layer to satisfy the above two constraints. Our network consumes a stack containing the reference image and two part mask images and outputs a feature vector. 
    We pair the feature vector with each vertex position $v_{i,j}$ of the two predicted part boxes and use another network to predict $\omega_{i,j}$'s. To make the contact point estimation invariant to the ordering of 3D bounding box vertices, the architecture of the second network is similar to the PointNet segmentation network~\cite{qi2017pointnet}. 
    Please refer to  Sec~\ref{sec:supp_relative_position} of the supplementary material for details.
    Qualitative results of relative position prediction are provided in supplementary material Fig~\ref{fig:center_module}. 
\end{itemize}
\section{Experiments}
We conduct experiments on PartNet~\cite{mo2019partnet} and compare our method with GenRe~\cite{zhang2018learning} and two other baselines that predict structured shapes.
Our method achieves superior performance over baselines on novel categories.
Ablation studies are included to validate our design of key modules.

\subsection{Dataset and Settings}
We use PartNet as the main testbed.
It provides fine-grained object part segmentation for 26,671 3D models, covering 24 object categories in ShapeNet~\cite{chang2015shapenet}.
We pick four most commonly used object categories: chair, table, cabinet and bed.
For all experiments, our method as well as baselines, are trained on the chair category only and tested on the rest three categories.

To generate the input images for training, we randomly sample 12 flying views and randomly move the shape location for each shape.
Our training set contains 2,576 chairs and 60,984 rendered images in total.
For the test set, we render each shape with 6 randomly sampled flying views.
The rendered image is of 256 $\times$ 256 resolution.

\subsection{Training Details}
Our proposed method contains several individual modules. At the training stage, we train each module separately. For example, to train the joint-based relative position module, our module takes the orientation and size from ground truth labels. For testing, we directly combine these modules. Empirically, we find our pipeline good enough even without further end-to-end joint fine-tuning.

\subsection{Baseline Methods}
\new{In addition to GenRe~\cite{zhang2018learning}, we provide two other baseline methods on predicting structured shapes: a naive encoder-decoder baseline and a graph convolutional network (GCN) baseline.  Note that there are no previous works addressing the exact same settings as ours. We try our best to adapt two previous works, namely StructureNet~\cite{mo2019structurenet} and Li~et~al.~\cite{li2020Learning3DPart}, to create the two baselines.}

\mypara{Naive Encoder-decoder Baseline.}
\new{This baseline borrows the network design from StructureNet~\cite{mo2019structurenet}. However, StructureNet assumes a shared canonical semantic part hierarchy, which is not suitable for cross-category generalization.  Thus, we remove the hierarchical part decoding stage in StructureNet and directly decode 50 part bounding boxes.}

\mypara{GCN Baseline.}
\new{For this baseline, we use a network architecture similar to  Li~et~al.~\cite{li2020Learning3DPart}, which builds part-graphs and uses the GCN to propagate information among adjacent part pairs. Concretely, for each part, we combine RGB input images and one additional channel for a 2D part mask to form a 4-channel input.  To train the GCN, we use the ground truth part masks and adjacent edges. }

\subsection{Evaluation Metric}
To quantitatively evaluate the reconstruction results, we use the Earth-Mover Distance (EMD)~\cite{Fan:2017}. It is a widely-used metric for comparing shape quality in 3D space.
To measure the shape distance against the GT point cloud, we generate a point cloud based on our part bounding box representation.
First, 1024 points are uniformly sampled from each estimated part bounding box.
Then, 1024 points are sampled from the union of sampled part points in the previous step by the farthest point sampling.


\subsection{Results and Analysis}
In this section, we provide both qualitative and quantitative comparisons with baseline methods. Table~\ref{tab:ablate_gnn} and Table~\ref{tab:res} compare our results to the GCN baseline and the naive encoder-decoder baseline respectively. We observe that all methods work well on the training category -- chair, while our method achieves significantly better results 
on novel test categories -- table, cabinet, and bed. 

\begin{figure*}[t]
\centering
\includegraphics[width=0.76\textwidth]{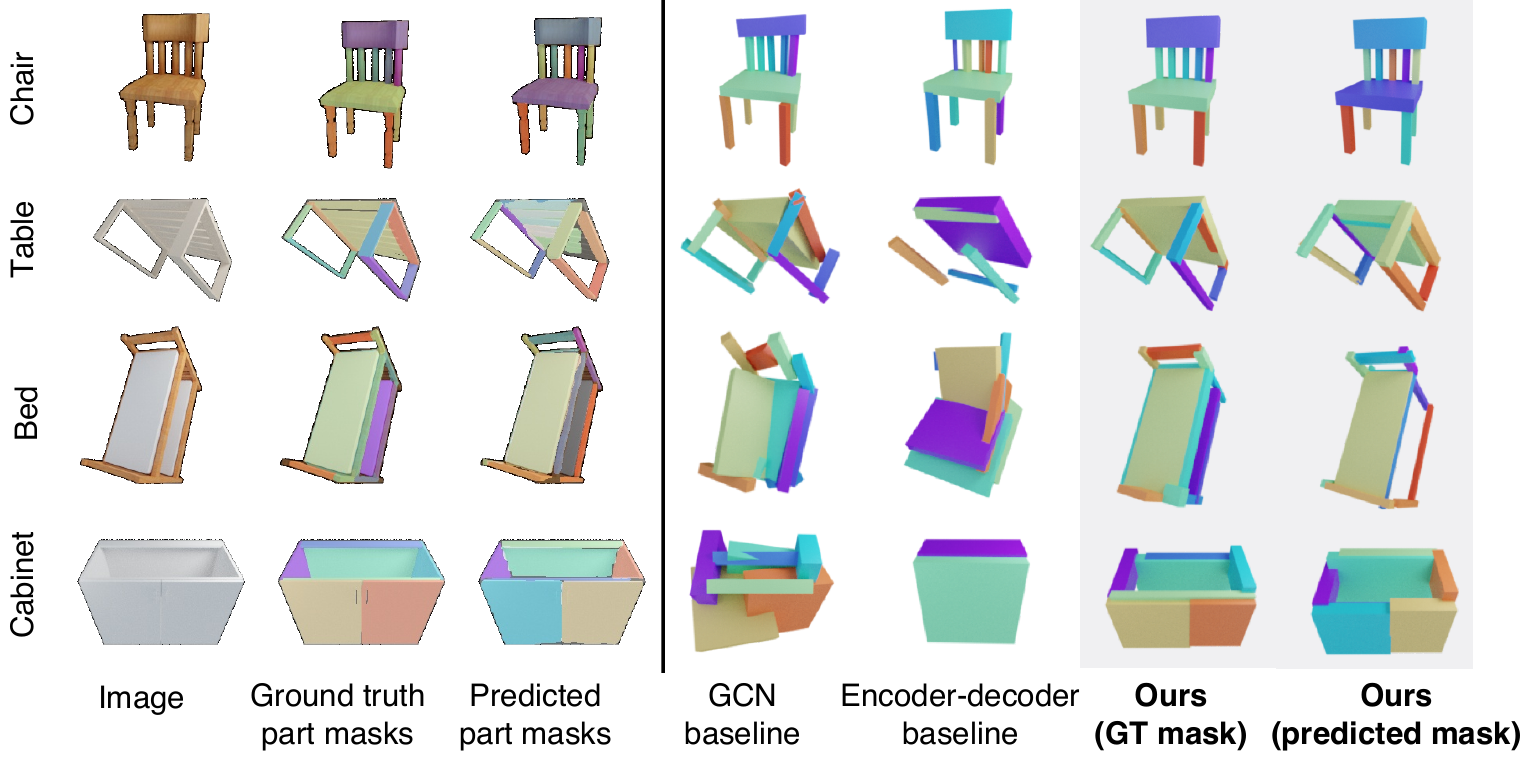}
\caption{\textbf{Qualitative comparisons to GCN baseline and Encoder-decoder baseline.} We visualize our 2D part mask predictions and the 3D part cuboids predictions. Notice that all models are trained on chairs only. The bottom three rows show shape reconstruction from novel unseen test categories: bed, cabinet, and table. 
}
\label{fig:dp-example}
\end{figure*}

\new{Furthermore, the comparison with the GCN baseline demonstrates the effectiveness of our modular pipeline.
For the comparison with the GCN, both the baseline and our method take the ground truth part mask and part adjacent relationship. Even with the ground truth part masks, the GCN fails to generalize to a novel category, which implies that explicit part inputs can not guarantee the compositional generalizability. With our carefully designed modules that encourage focusing on local structures, significantly better results are achieved in novel categories. We illustrate this finding in a qualitative comparison in Fig.~\ref{fig:dp-example}. On the chair example, baseline methods achieve very competitive results compared to our method. However, the two baseline methods fail to generate satisfying results on novel categories (tables, beds, and cabinets), while our method is still able to do high-fidelity structure prediction.}

\begin{figure*}[t]
\centering
\includegraphics[width=0.76\textwidth]{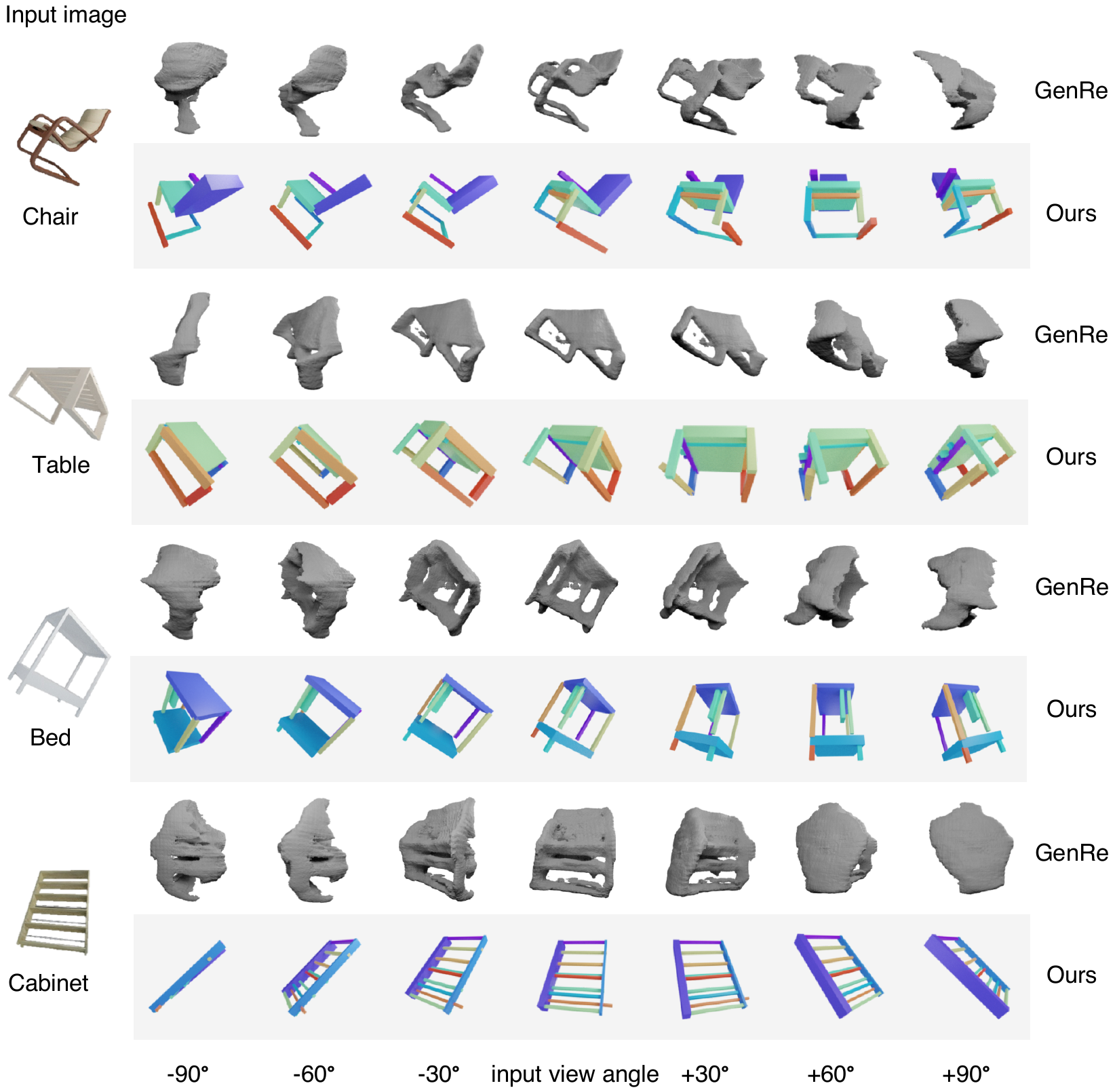}
\caption{\textbf{Qualitative comparisons to GenRe~\cite{zhang2018learning}.} We compare our results with predicted mask to GenRe and present each result from multiple views to clearly show the 3D reconstruction. While the results from GenRe look good from the input image views (the fifth column), the 3D reconstruction of GenRe is actually worse than ours as compared from other views.
}
\label{fig:compare_genre}
\end{figure*}

\new{Besides, we provide a qualitative (Fig.~\ref{fig:compare_genre}) and quantitative (Table~\ref{tab:comp_genre}) comparison to GenRe~\cite{zhang2018learning}. Note GenRe has a different setting from our method. GenRe outputs a mesh surface, while we output the part bounding boxes.
Nonetheless, we provide comparison based on EMD in Table~\ref{tab:comp_genre} using point cloud sampled from surfaces, which do not favor our structured part bounding boxes. }
{Thanks to the structure prior, 
our method outperforms GenRe by a large margin.  
The qualitative comparison to GenRe in Fig.~\ref{fig:compare_genre} shows that our method recovers the shape structure more faithfully.}

\begin{table}[t]
\centering
\small
\begin{tabular}{@{}lllll@{}}
\hline
 & Chair $\downarrow$ & Table $\downarrow$ & Cabinet $\downarrow$ & Bed $\downarrow$ \\ \hline
GCN & 0.026	& 0.073 & 0.080 & 0.053 \\
Ours (GT mask) & \textbf{0.017}  & \textbf{0.048} & \textbf{0.053}  & \textbf{0.033} \\ \hline
\end{tabular}
\caption{\textbf{Quantitative comparisons with the GCN baseline.} (EMD on point clouds sampled from 
part bounding boxes.)
}
\label{tab:ablate_gnn}
\vspace{-2mm}
\end{table}

\begin{table}[t]
\centering
\small
\begin{tabular}{@{}lllll@{}}
\hline
 & Chair $\downarrow$ & Table $\downarrow$ & Cabinet $\downarrow$ & Bed $\downarrow$ \\ \hline
Naive & \textbf{0.026}&  0.091&  0.095&  0.057  \\
Ours (predicted mask) & 0.028 & \textbf{0.054} & \textbf{0.060} & \textbf{0.042} \\ \hline
\end{tabular}
\caption{\textbf{Quantitative comparison with the naive encoder-decoder baseline.} (EMD on point clouds sampled from part bounding boxes.)
} 
\label{tab:res}
\end{table}

\begin{table}[t]
\centering
\small
\begin{tabular}{@{}lllll@{}}
\hline
      & Chair $\downarrow$ & Table $\downarrow$ & Cabinet $\downarrow$ & Bed $\downarrow$ \\ \hline
GenRe~\cite{zhang2018learning} & 0.056 & 0.099 & 0.081 &  0.075  \\
Ours (predicted mask)  & \textbf{0.032} & \textbf{0.053} & \textbf{0.053} & \textbf{0.041} \\ \hline
\end{tabular}
\caption{\textbf{Quantitative comparison with GenRe~\cite{zhang2018learning}.} (\textit{EMD on surface point clouds.})}
\label{tab:comp_genre}
\end{table}

\subsection{Ablation Studies}
\label{sec:exp:ablation}
We conduct several ablation experiments to demonstrate the effectiveness of our key components.

\mypara{Part Orientation Prediction.}
\new{Table~\ref{tab:ablate_rot} compares different losses w.r.t the average geodesic error between GT and predicted rotation matrices.
\emph{MSE} stands for mean squared error, which does not consider rotation ambiguity. It performs much worse than other losses, which indicates that rotation ambiguity hurts the learning process. With \emph{MinN-MSE}, we achieve significantly better rotation prediction. Besides, \emph{MOE-MinN-MSE} outperforms \emph{MinN-MSE} on all categories, indicating that multiple predictions facilitate the network to escape poor local minima.}
For \emph{MOE-Prior-MSE}, we use a heuristic algorithm to label part symmetry axes. Compared to our \emph{MOE-MinN-MSE}, the performance is worse.

\begin{table}[t]
\centering
\small
\begin{tabular}{@{}lllll@{}}
\hline
Loss & Chair $\downarrow$ & Table $\downarrow$ & Cabinet $\downarrow$ & Bed $\downarrow$ \\ \hline
MSE & 38.03 & 39.43	& 39.82 & 38.98 \\
MinN-MSE &  9.21 & 13.74 & 8.29 & 10.81 \\
MOE-Prior-MSE & 9.85 & 14.10 & 9.75 & 11.35\\
\textbf{MOE-MinN-MSE} & \textbf{7.11} & \textbf{10.31} & \textbf{3.42} & \textbf{6.43} \\
\hline
\end{tabular}
\caption{\textbf{Ablation experiments with MoE-MinN orientation prediction. }
Comparison of average geodesic errors (angular difference in degrees) between ground truth and predicted rotation matrices over different losses.}
\label{tab:ablate_rot}
\end{table}

\mypara{Group-based Size Prediction.}
\new{
Table~\ref{tab:ablate_group_size} compares unary and group-based size predictions.
Our group-based size prediction shows consistently superior performance, compared with the unary counterpart.
It implies that the group-based size prediction might benefit from information aggregated from multiple parts in the group. 
}

\begin{table}[t]
\centering
\begin{tabular}{lllll}
\hline
 & Chair $\downarrow$ & Table $\downarrow$ & Cabinet $\downarrow$ & Bed $\downarrow$ \\ \hline
unary size & 0.054 & 0.134 & 0.183 & 0.136 \\
group size & \textbf{0.050} & \textbf{0.115} & \textbf{0.175} & \textbf{0.130} \\ \hline
\end{tabular}
\caption{\textbf{Ablation experiments on group-based size prediction.} We report the number of average L1 distance.}
\label{tab:ablate_group_size}
\end{table}

\mypara{Joint-based Relative Position Prediction.}
Table~\ref{tab:ablate_center} compares our joint-based relative position prediction with absolute position prediction. Both approaches take images, part masks, part orientations, and part sizes as input.  We achieve much more accurate results across all categories. As absolute locations are sensitive to shape scales and translations, predicting absolute location is a more difficult problem than relative position.

\begin{table}[t]
\centering
\begin{tabular}{lllll}
\hline
 & Chair $\downarrow$ & Table $\downarrow$ & Cabinet $\downarrow$ & Bed $\downarrow$ \\ \hline
absolute & 0.035 & 0.080 & 0.082 & 0.067 \\
relative (ours) &  \textbf{0.023} & \textbf{0.038} & \textbf{0.045} & \textbf{0.042}  \\ 
\hline
\end{tabular}
\caption{\textbf{Ablation experiments on relative position prediction.} Comparison of our relative position prediction to absolute position prediction with L1 distance.}
\label{tab:ablate_center}
\end{table}

\subsection{Real World Data}
\begin{figure}
    \centering
    \includegraphics[width=\linewidth]{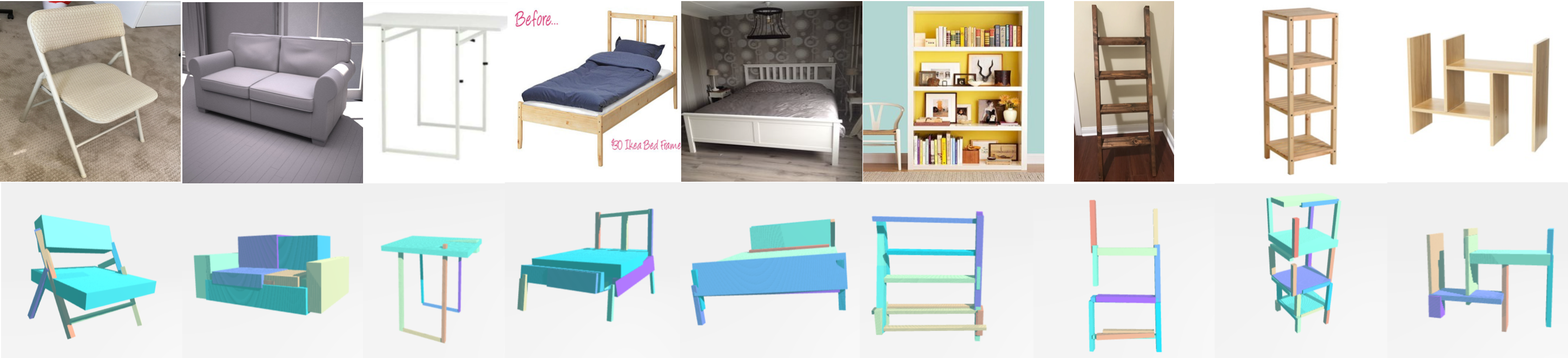}
    \caption{\textbf{Qualitative examples on real images} (best viewed zoom-in). Note that the object is masked out by the GT silhouette.}
    \label{fig:real_img_results}
\end{figure}

Examples of 3D reconstruction results on real world images are shown in Fig~\ref{fig:real_img_results}. We use images from Pix3D~\cite{pix3d} and the Internet.
The object is masked out by the ground truth silhouette.
Our model shows satisfactory results on real world images, even though only trained on synthetic ones.

\section{Conclusion}
In this paper, we first bring the concept of compositional generalizability to the task of estimating 3D shape structures from a single 2D image input and demonstrate much better reconstruction capabilities that generalize to the shapes from unseen object categories.
At the core of our design, we factorize the whole problem into sub-problems that explicitly discover the commonly shared part relationships and shape substructure priors across different object categories.
We carefully design the network module for each of the sub-problem, which uses localized contexts for the generalizability guarantee.
Experiments show that our method is effective on generalizing to novel categories.

{\small
\bibliographystyle{ieee_fullname}
\bibliography{arxiv}
}

\newpage
\appendix
\noindent
{\Large \textbf{Supplementary Materials}}
\appendix

{
  \etocdepthtag.toc{mtappendix}
  \etocsettagdepth{mtchapter}{none}
  \etocsettagdepth{mtappendix}{subsection}
  \hypersetup{linkcolor=blue}
  \tableofcontents
}

\section{Overview}
This document accompanies the main paper and provides additional qualitative results for the entire pipeline, extensive evaluations for each individual sub-module in our pipeline, elaborations on the network architecture designs, and 
more ablation studies. 
We will release our code, data, and pre-trained models upon paper acceptance.

Sec.~\ref{sec:more-results} provides more qualitative results of our method and comparisons to the three baseline methods.
%
In Sec.~\ref{sec:individual-module-results}, we provide more detailed evaluations of each individual network module.
%
Sec.~\ref{sec:impl-details} presents more details about our network architecture designs and the implementation.
Sec.~\ref{sec:ablation} includes more ablation studies, besides the three shown in the main paper, to further analyze the effectiveness of our design choices in each module. 

\section{More Qualitative Results}
\label{sec:more-results}
\begin{figure*}
    \centering
    \includegraphics[width=\textwidth]{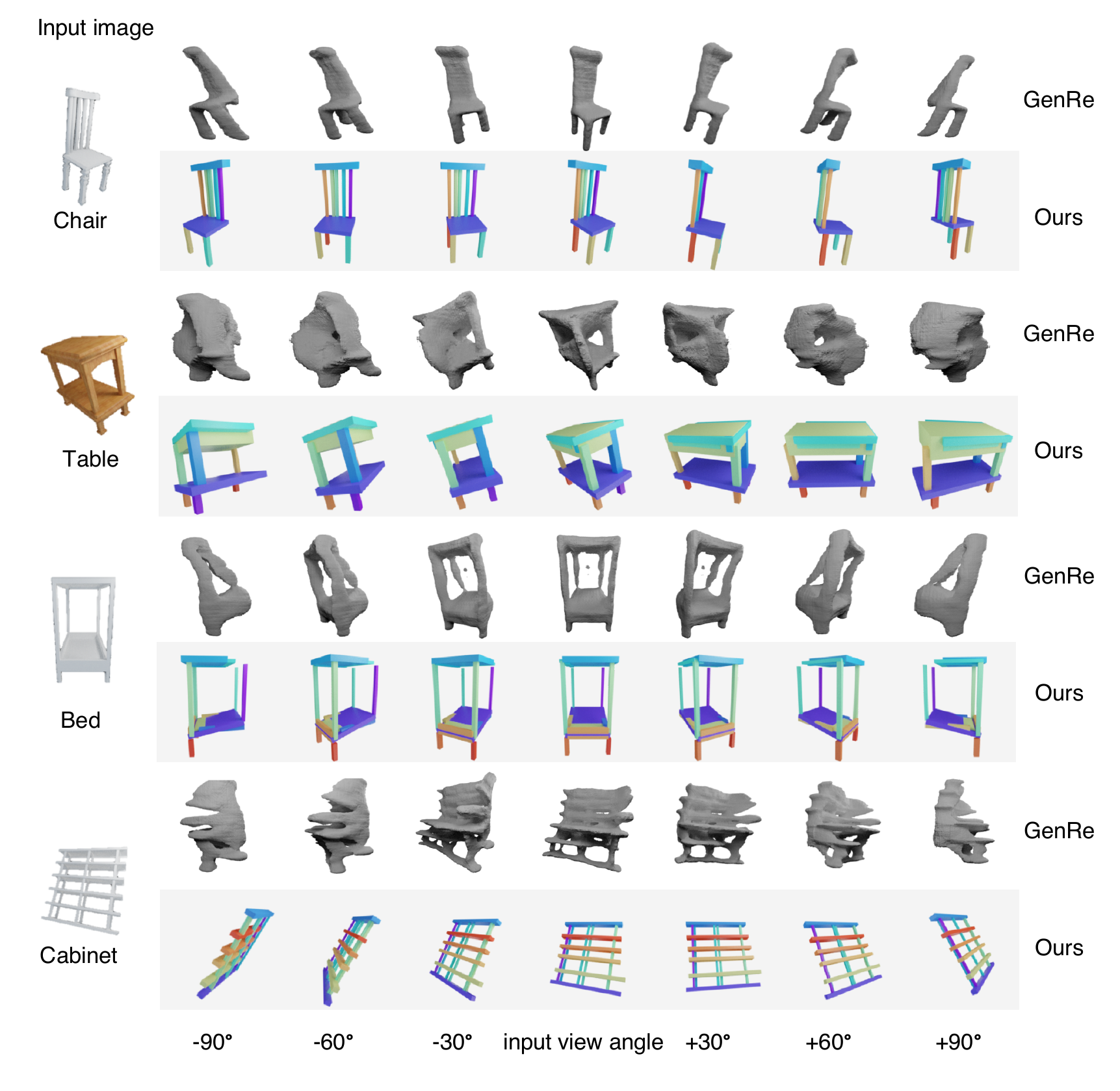}
    \caption{\titlecap{More Comparisons to GenRe~\cite{zhang2018learning}.}{We compare our results to GenRe and present each result from multiple views to clearly show the 3D reconstruction. 
    For each object category, we present three rows of results for GenRe, ours(GT mask), ours(predicted mask) respectively, with the first column shows the input image.
    While the result from GenRe seems to be good from the input image views (the fifth column), the 3D reconstruction of GenRe is actually worse than ours if we compare from the other views.
    We also see that the GenRe model trained on chairs tends to overfit to the chair shapes, while our method can faithfully reconstruct shapes from unseen test categories.
    }}
    \label{fig:multiview_genre}
\end{figure*}

We show more qualitative results comparing our methods against the naive encoder-decoder baseline and the graph-convolution-network baseline.
Figures~\ref{fig:ours_chair}, \ref{fig:ours_table}, \ref{fig:ours_bed}, and \ref{fig:ours_cabinet} show more chair, table, bed, and cabinet results respectively.
Notice that all methods are trained on chairs only.
We see that our method achieves much better reconstruction generalizability than the baseline methods.
We also present more qualitative comparisons to GenRe~\cite{zhang2018learning} in Fig.~\ref{fig:multiview_genre}, where we can clearly see that our method reconstructs the shape structure more faithfully.

\section{Evaluations of Individual Modules}
\label{sec:individual-module-results}
Our pipeline has many sub-network modules.
Due to the page limit of the main paper, we cannot present detailed result visualization and discussion on each sub-network module.
In this section, we provide qualitative and quantitative evaluations for several key network modules to illustrate the predicted results at each step in our pipeline.

Concretely, we measure the quantitative accuracy of the part instance segmentation module and part adjacency relationship module.
Then, we show qualitative results and analysis on the part orientation module, the part size module, and the part relative position module. 

\subsection{Part Instance Segmentation}

\begin{figure*}[t]
    \centering
    \includegraphics[width=\textwidth]{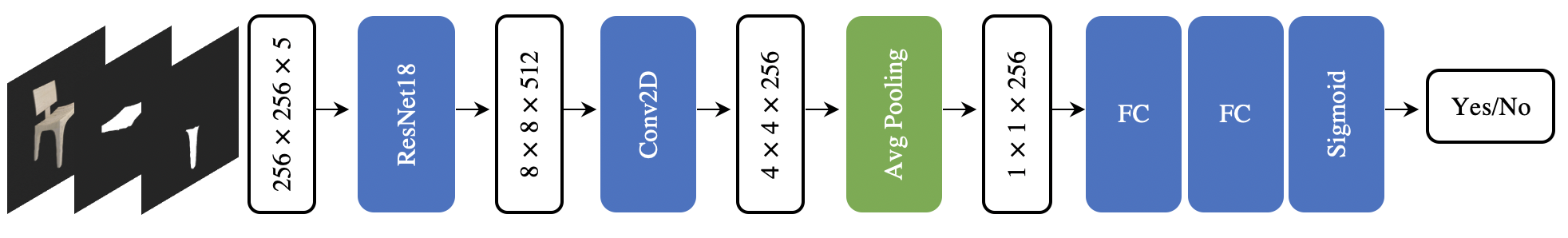}
    \caption{\titlecap{Network Architecture of the Part Relationship Prediction Module.}{This module is formulated as a binary classification network, which detects translational symmetry and adjacency of two parts from the image and their masks. Our part relationship module takes one image and two part masks as inputs, extract image global feature with ResNet18 without pre-training, a convolution layer, and an average-pooling layer. The global feature is then forwarded to two fully-connected layers and activated by a Sigmoid function. The output is the probability of the existence of a certain part relationship.}}
    \label{fig:relation}
\end{figure*}

\begin{figure*}
    \centering
    \includegraphics[width=0.9\textwidth]{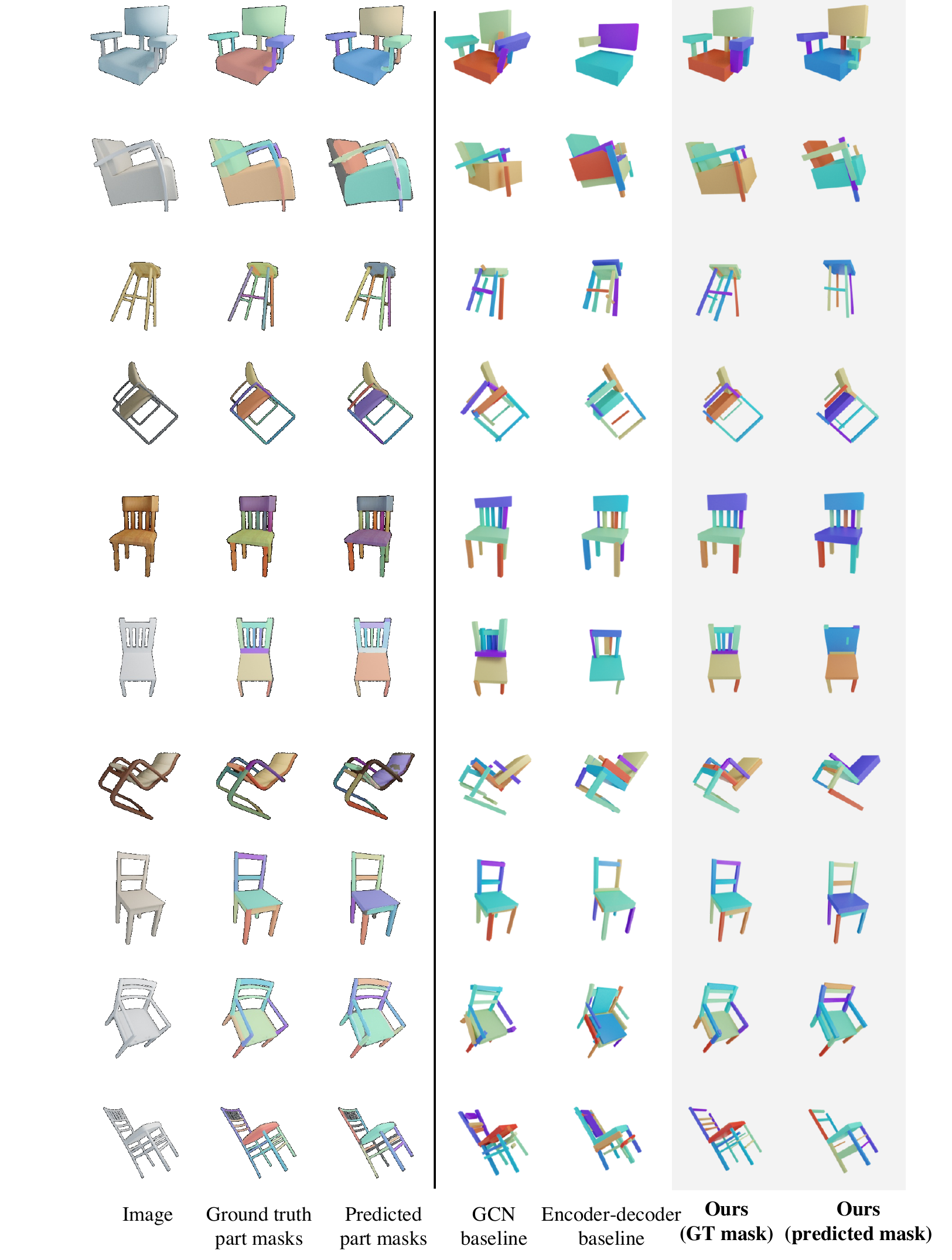}
    \caption{\titlecap{More Qualitative Results and Comparisons on Chairs (training category).}{From left to right, we present: input 2D image, GT 2D part mask, predicted 2D part mask, and 3D predictions of the GCN baseline, the naive encoder-decoder baseline, ours(GT mask), ours(predicted mask).}}
    \label{fig:ours_chair}
\end{figure*}

\begin{figure*}
    \centering
    \includegraphics[width=0.9\textwidth]{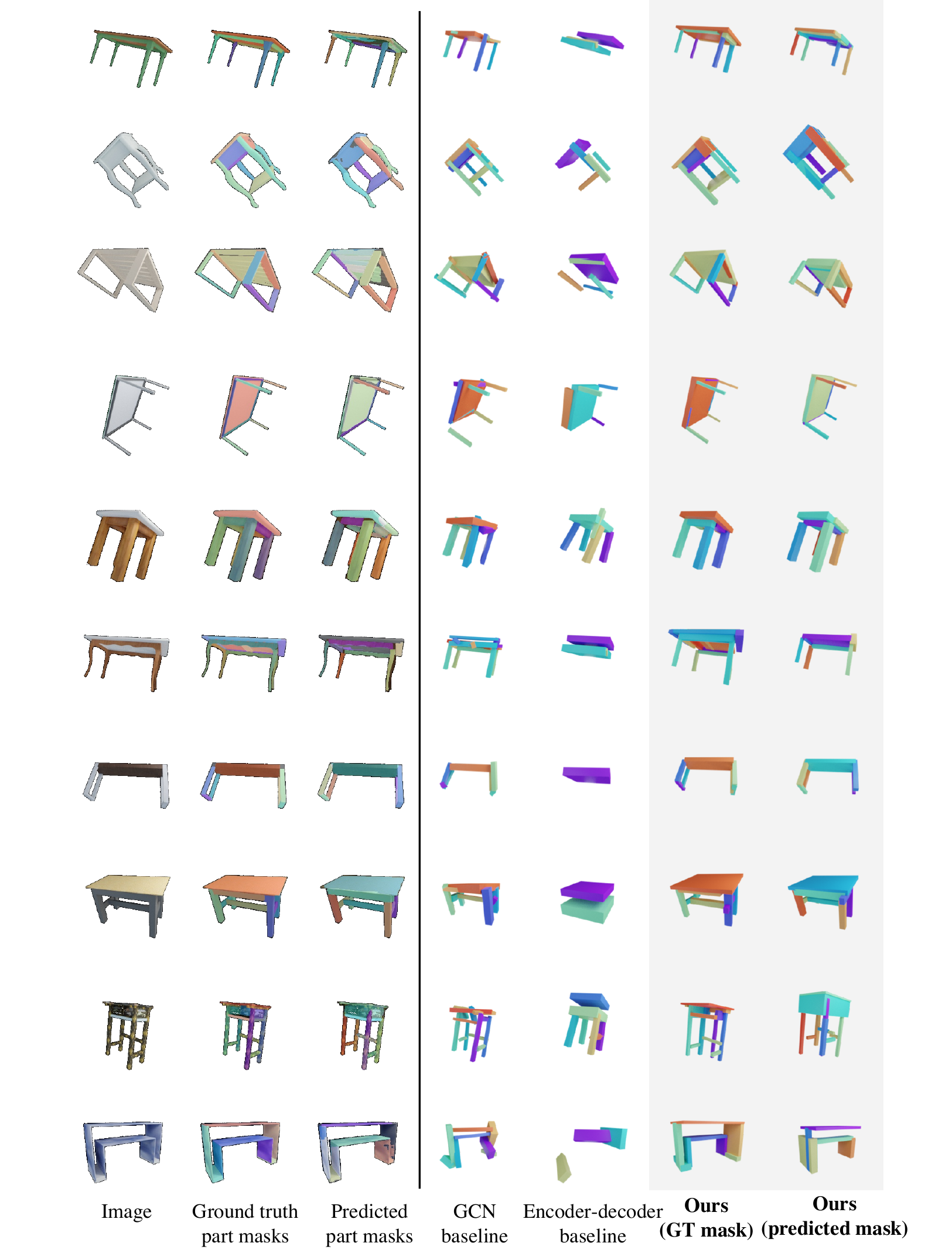}
    \caption{\titlecap{More Qualitative Results and Comparisons on Tables (novel category).}{From left to right, we present: input 2D image, GT 2D part mask, predicted 2D part mask, and 3D predictions of the GCN baseline, the naive encoder-decoder baseline, ours(GT mask), ours(predicted mask).}}
    \label{fig:ours_table}
\end{figure*}

\begin{figure*}
    \centering
    \includegraphics[width=0.9\textwidth]{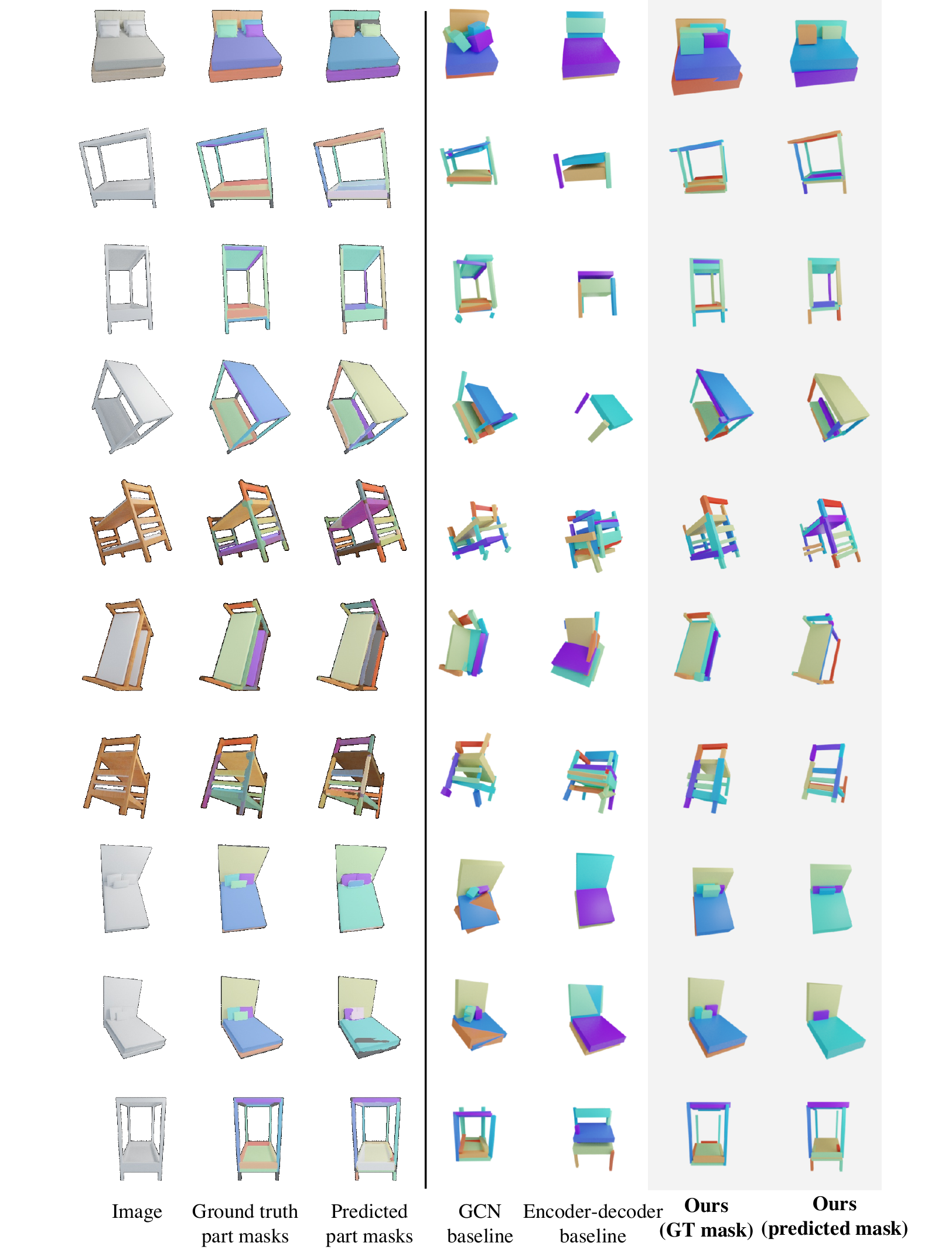}
    \caption{\titlecap{More Qualitative Results and Comparisons on Beds (novel category).}{From left to right, we present: input 2D image, GT 2D part mask, predicted 2D part mask, and 3D predictions of the GCN baseline, the naive encoder-decoder baseline, ours(GT mask), ours(predicted mask).}}
    \label{fig:ours_bed}
\end{figure*}

\begin{figure*}
    \centering
    \includegraphics[width=0.9\textwidth]{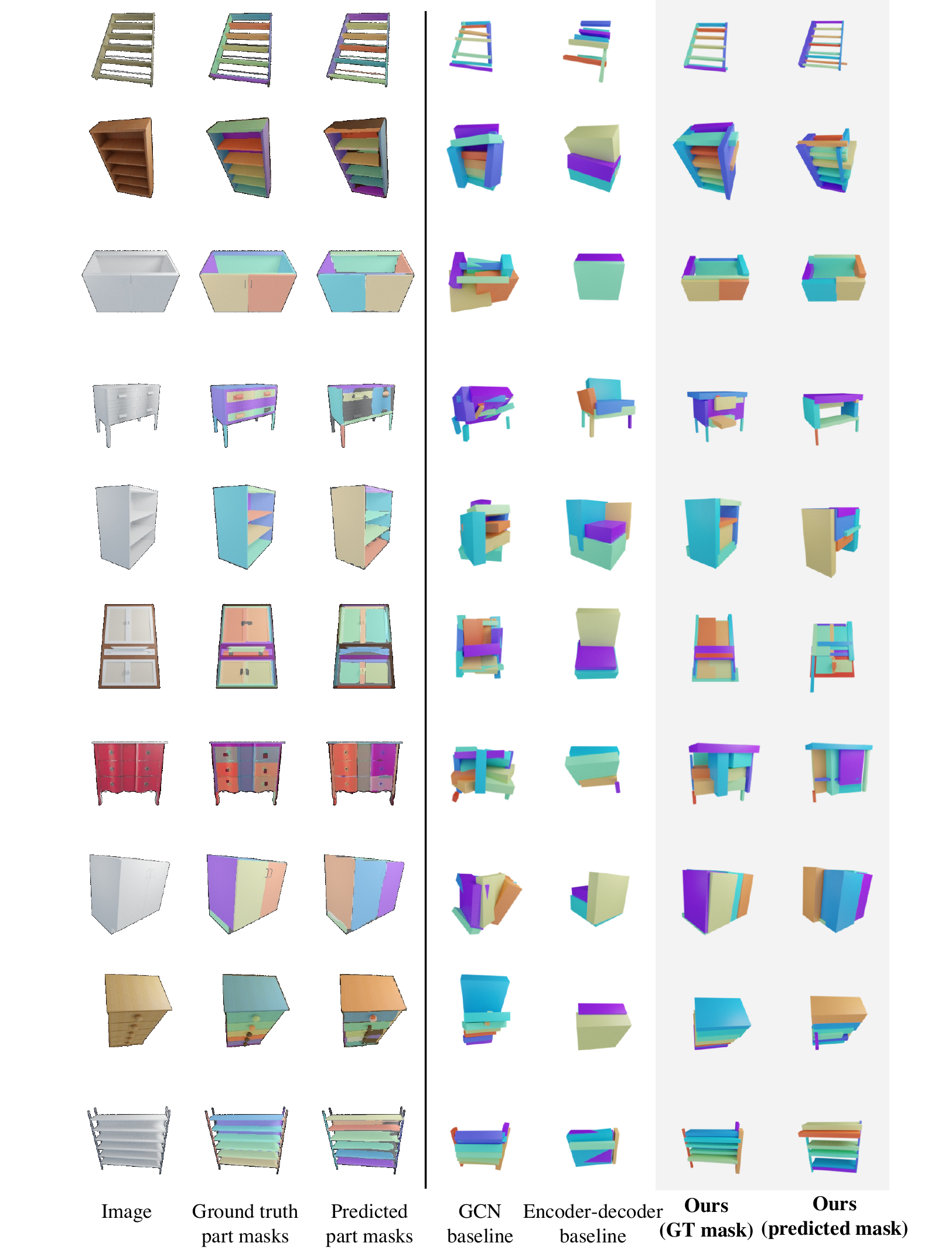}
    \caption{\titlecap{More Qualitative Results and Comparisons on Cabinets (novel category).}{From left to right, we present: input 2D image, GT 2D part mask, predicted 2D part mask, and 3D predictions of the GCN baseline, the naive encoder-decoder baseline, ours(GT mask), ours(predicted mask).}}
    \label{fig:ours_cabinet}
\end{figure*}

In Table~\ref{tab:part_mask}, we evaluate our 2D part mask instance segmentation module using the standard 2D object detection metrics: bounding box average precision (AP) and mask AP.
We find that, though trained on chairs only, our part instance segmentation module generalizes well to novel categories in a zero-shot fashion.
Figures~\ref{fig:ours_chair}, \ref{fig:ours_table}, \ref{fig:ours_bed}, and \ref{fig:ours_cabinet} (left three columns) visualizes example predictions on the four object categories: chairs (training category), beds, tables, and cabinets (test categories).
We can clearly see that the 2D part mask predictions are accurate enough to support the later 3D reasoning stages of our pipeline.

\begin{figure*}[t]
    \centering
    \includegraphics[width=\textwidth]{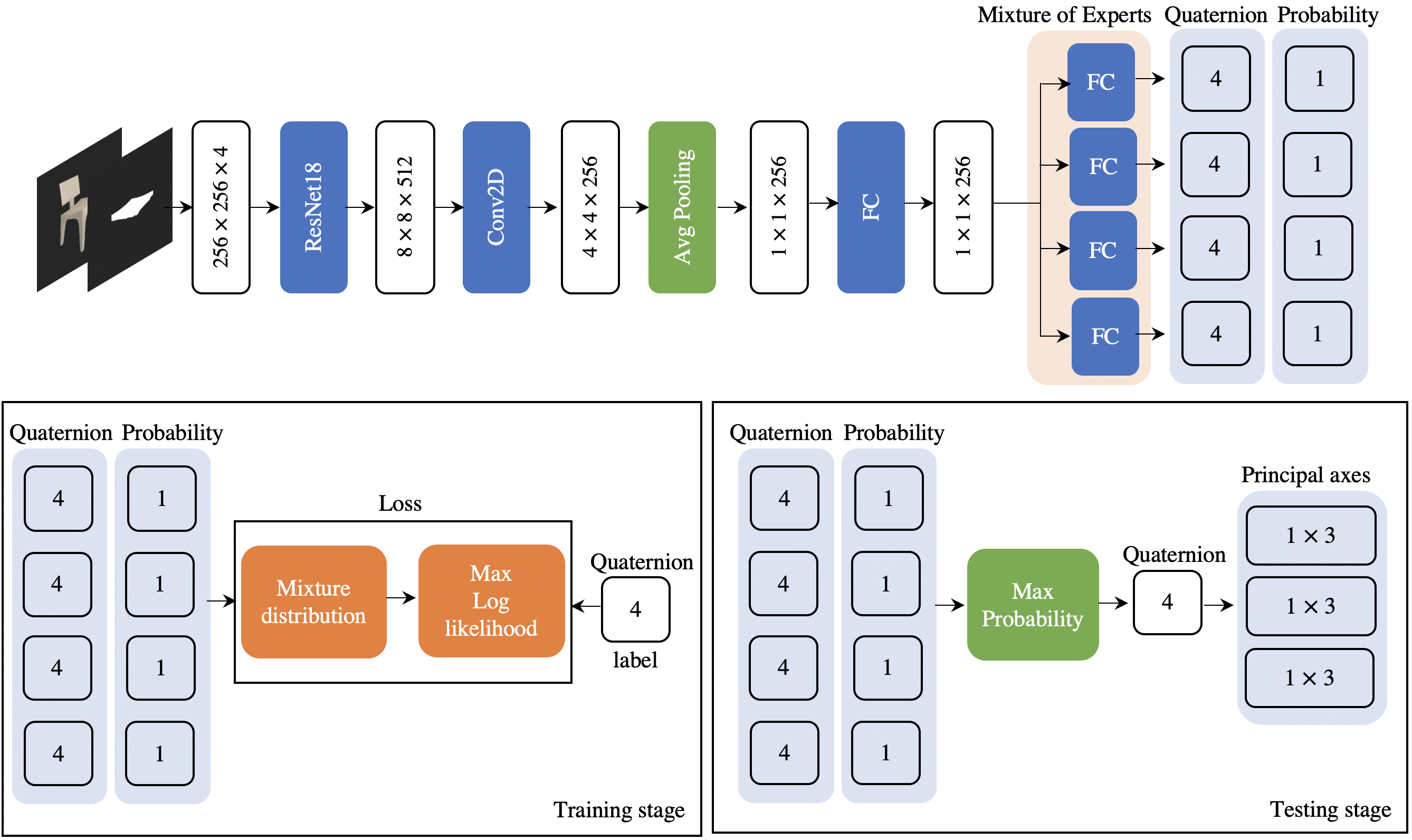}
    \caption{\titlecap{Network Architecture of the Part Orientation Prediction Module.}{This network takes one image and one part mask as input, and extracts a global image feature through a ResNet18, followed by a convolution layer and an average pooling layer. Then, we regress the quaternion outputs from the global feature with fully connected layers. We predict 4 rotations as well as the corresponding selection probability for the proposed Mixture-of-Experts approach. 
    At training stage, we maximize the log likelihood under the rotation distribution.
    At testing stage, we pick the rotation with the maximal probability.
    }}
    \label{fig:unary_rotation}
\end{figure*}

\begin{table}[t]
\centering
\begin{tabular}{lllll}
\toprule
 & Chair & Table & Cabinet & Bed \\ \midrule
box AP@0.5 & 83.18 & 72.03 & 34.40 & 55.31 \\ 
mask AP@0.5 & 79.11 & 66.53 & 25.08 & 47.89 \\ 
\bottomrule
\end{tabular}
\caption{\titlecap{Results of Our Part Instance Segmentation.}{The metrics include bounding box AP (average precision) and mask AP with the IoU threshold 0.5.}}
\label{tab:part_mask}
\end{table}

\subsection{Part Orientation Prediction}
\begin{figure*}
    \centering
    \includegraphics[width=1.0\textwidth]{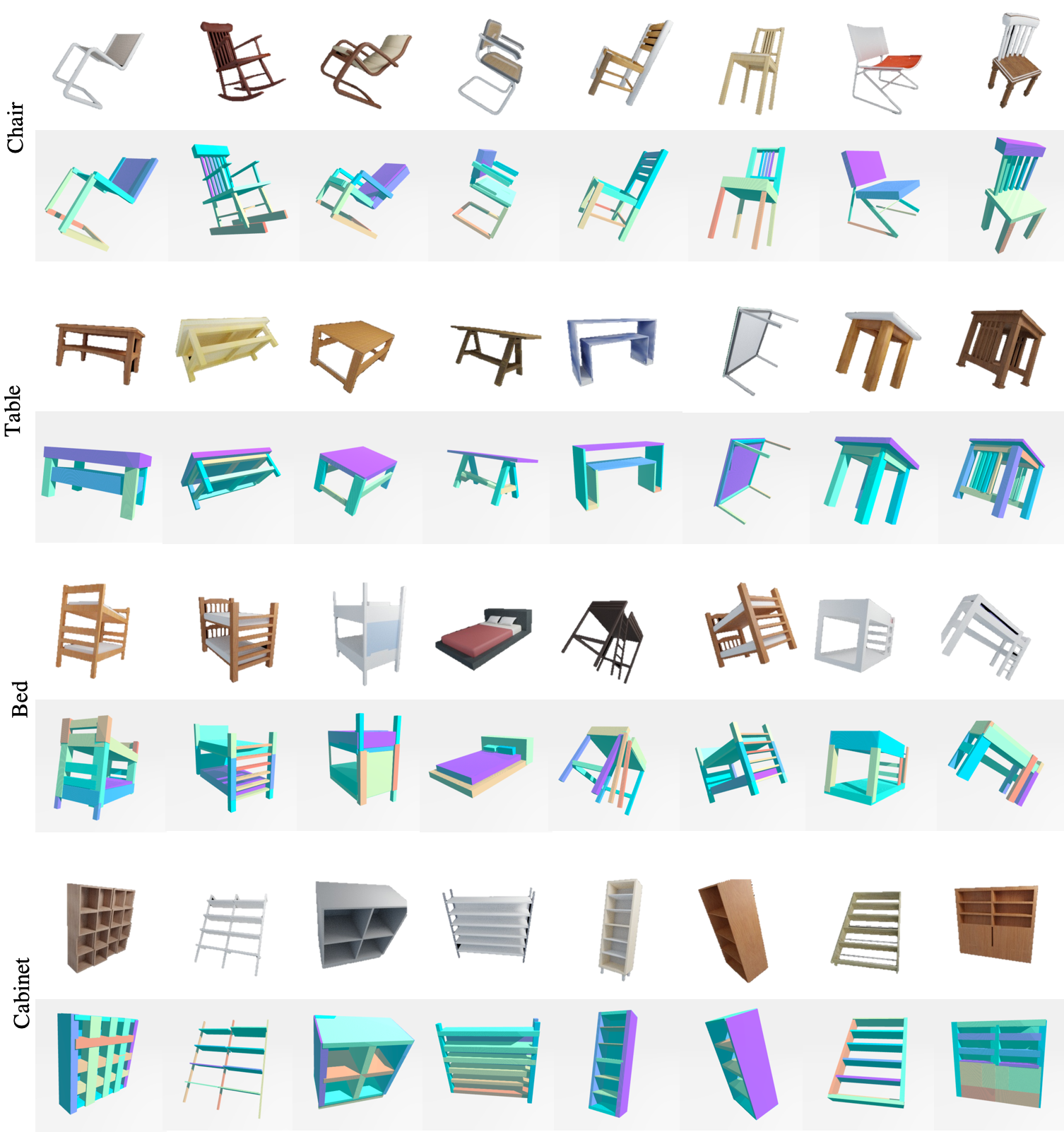}
    \caption{\titlecap{Part Orientation Predictions.}{For each part, we use the estimated orientation and the ground-truth size and center position. For each object category, the first row shows the input images and the second row presents the 3D part bounding boxes reconstructions.}}
    \label{fig:rot_module}
\end{figure*}

Fig.~\ref{fig:rot_module} shows qualitative results of our predicted part orientation. To visualize each part oriented bounding box (OBB), we take the estimated part rotation and use ground truth part size and position. 
{As our orientation module predicts principal axes of OBB in an orderless fashion, we re-order the ground truth length along each axis based on the predicted axes and ground truth axes. }
While our model is only trained on chairs, we achieve superior results on table, bed, and cabinet categories.

\subsection{Part Size Prediction}
\begin{figure*}
    \centering
    \includegraphics[width=1.0\textwidth]{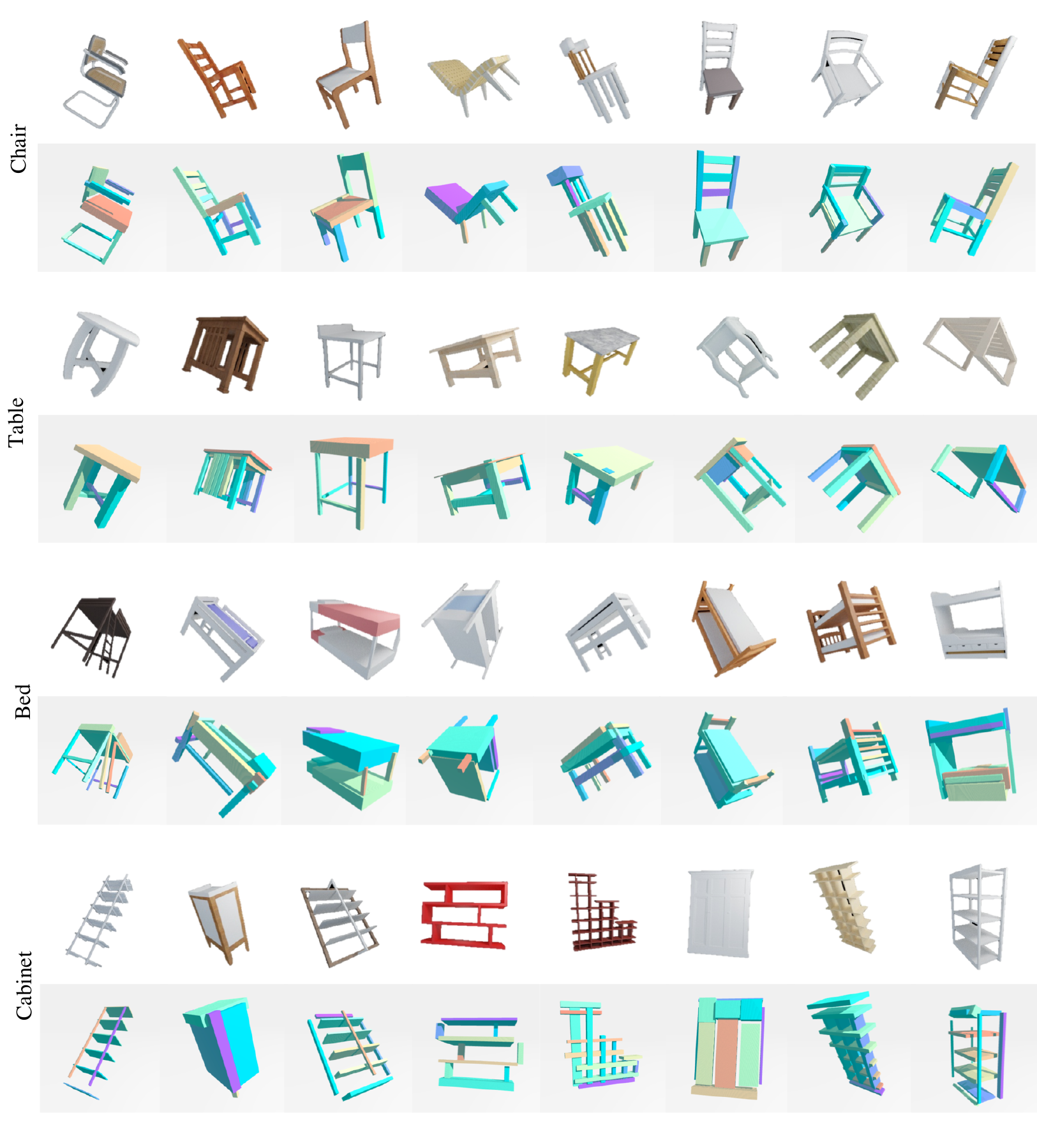}
    \caption{\titlecap{Group-based Part Size Predictions.}{To visualize each part, we use the estimated size and the ground truth part orientation and part position. For each object category, the first row shows the input images and the second row presents the 3D part bounding boxes reconstructions. Note that results may show disconnected parts since the part 3D box use the ground truth center positions. Our follow-up shape assembly process will connect adjacent parts.}}
    \label{fig:size_module}
\end{figure*}

Fig.~\ref{fig:size_module} presents qualitative example results of our group-based size prediction. To evaluate solely the performance of the group-based size module, we take ground truth orientation as input. We visualize each part with the estimated size and the ground truth orientation and position. High quality results on table, bed and cabinet demonstrate that our part size module can generalize well to the unseen test categories.

\subsection{Part Adjacency Relationship Prediction}

\begin{table}[t]
\centering
\begin{tabular}{lllll}
\toprule
 & Chair & Table & Cabinet & Bed \\ \midrule
Accuracy & 0.943 &  0.907 & 0.913 & 0.873 \\ \bottomrule
\end{tabular}
\caption{\titlecap{Prediction Accuracy of Estimated Part Adjacency Relationships.}{We see that the part adjacency relationships can be easily learned and generalized to unseen test categories}}
\label{tab:part_adj}
\end{table}

Our shape assembly process depends on a part-graph structure to form a sequential part assembly tree.
We leverage part adjacency relationships as the pairwise edges connecting the neighboring parts  in the part-graph.
We train a network that predicts whether two parts are adjacent or not to construct such part-graph structure.
Table~\ref{tab:part_adj} shows the quantitative evaluation of our part adjacency relationship predictions, where we see that the part adjacency relationships can be easily learned and generalized to unseen test categories.

\subsection{Shape Assembly Process}
\begin{figure*}
    \centering
    \includegraphics[width=1.0\textwidth]{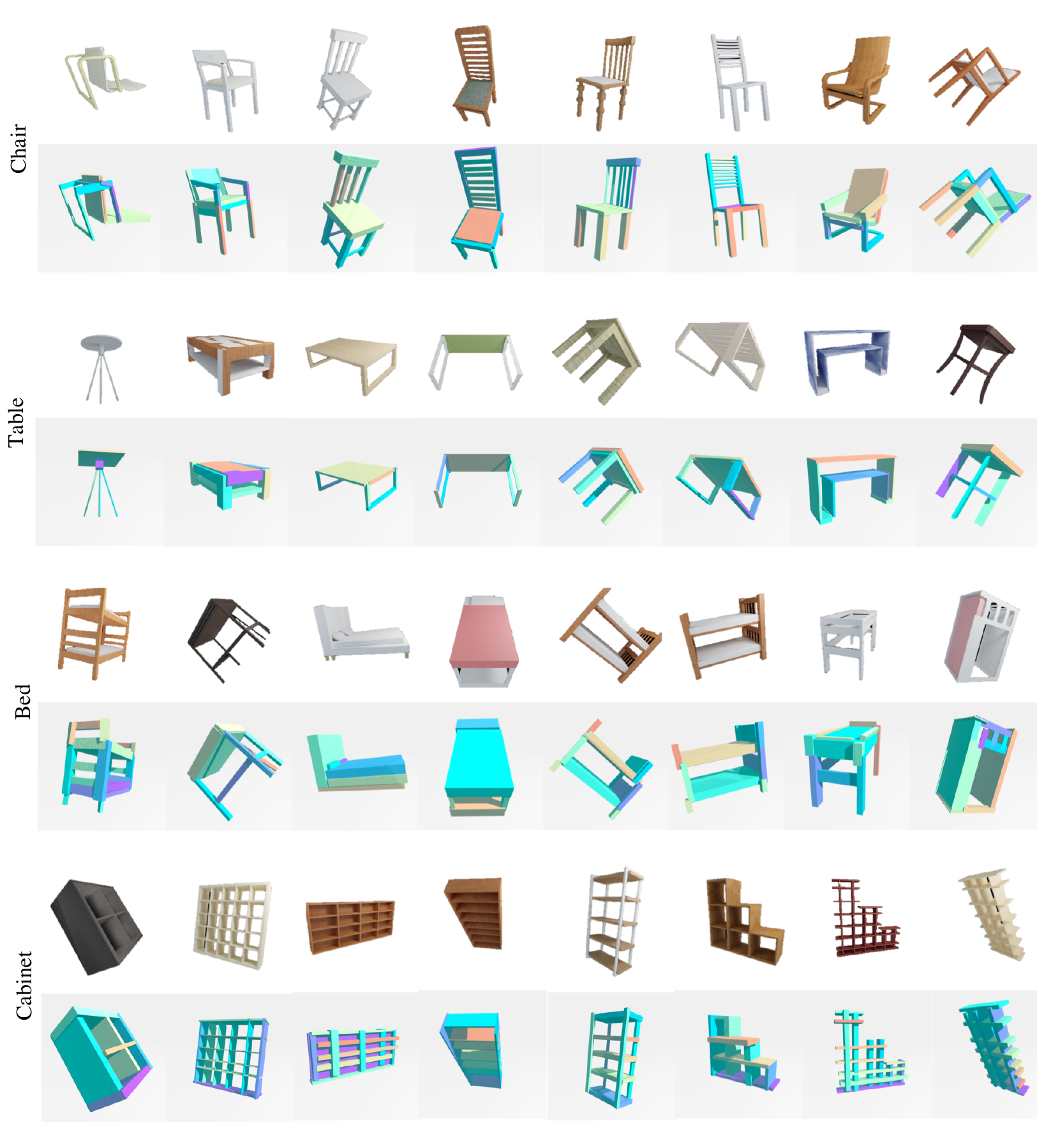}
    \caption{\titlecap{Joint-based Part Relative Position Predictions.}{To visualize each part, we use the estimated center position and the ground truth part orientation and length. For each object category, the first row shows the input images and the second row presents the 3D part box reconstructions.}}
    \label{fig:center_module}
\end{figure*}

Fig.~\ref{fig:center_module} shows some example results for our joint-based part relative position estimation. The results here take the ground truth part orientation and size as input. Each part is visualized with the estimated position and the ground-truth orientation and size. Though our model is trained on chairs only, our position estimation module can work well on very different shapes from the cabinet, bed, and table categories.

\section{Network Architecture Details}
\label{sec:impl-details}
In this section, we describe more network architecture and implementation details for our network modules.

\subsection{Part Instance Segmentation}
\label{sec:supp_part_instance_seg}
We use Mask-RCNN~\cite{he2017mask} implemented in Detectron2~\cite{wu2019detectron2} to do part instance segmentation, which is originally designed for COCO~\cite{lin2014microsoft} object instance segmentation.
Since the aspect ratios of PartNet~\cite{mo2019partnet} parts follow a different distribution from those of COCO objects, we add two additional aspect ratios (1:4 and 4:1) to each anchor (predefined sliding windows).
The input image is resized to $800 \times 800$, and flipped randomly as data augmentation.
The detector is a ResNet50-based FPN, pretrained on COCO.
It is fine-tuned on the chair category for 80,000 iterations by SGD.
The batch size is 4.
The initial learning rate is 0.005, and is divided by 10 after 48,000 and 64,000 iterations.
Other hyperparameters follow the default setting of Detectron2.

\subsection{Part Relationship Prediction}
\label{sec:supp_part_relation}
In our method, we detect translational symmetry and adjacency part relationships.
The detection of relationship is modeled as a binary classification problem.
The architecture of the relationship detection module is visualized in Fig.~\ref{fig:relation}.
The input to the network is the concatenation of an image and two part masks.
A ResNet18 without pre-training followed by a convolution layer and an average-pooling layer is used to extract a global feature.
The global feature is forwarded to two fully-connected layers and activated by a sigmoid function.
The final output is the probability of the existence of a certain relationship between input parts. In our experiments, we use threshold 0.7 for adjacency relationship and threshold 0.9 for translation symmetry relationship.
The module is trained on the chair category for 10 epochs by the Adam optimizer.
The batch size is 64.
The initial learning rate is 0.001, and decays by 0.9 every 2 epochs.

\begin{figure*}[t]
    \centering
    \includegraphics[width=\textwidth]{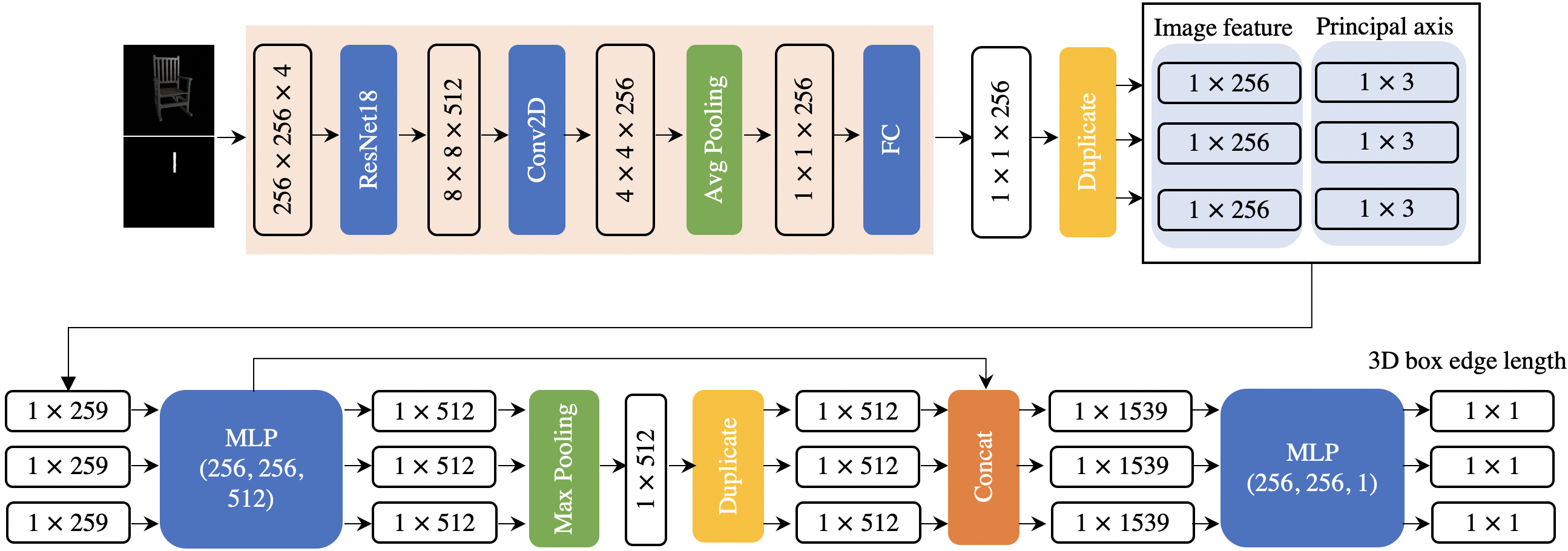}
    \caption{\titlecap{Network Architecture of the Single Part Size Prediction Module.}{ This network takes an image, a part mask, and part principal axes (orientation) as input. The goal is to predict the edge length of part bounding box along each axis.
    Our size prediction module adopts a similar architecture to the PointNet segmentation network to maintain the property that edge lengths should be permutation equivariant to the order of principal axes. We consider each axis a ‘point’, and estimate its attribute (edge length) in a point-wise segmentation fashion. we first concatenate the image feature and each axis as the input. Then, we apply a shared MLP to extract the feature of each axis.  Following PointNet, a global feature is also extracted through the share MLP and concatenated with each local feature to enable the information communication between different axes. Finally, a scalar edge length is estimated from the feature of each axis.
    }}
    \label{fig:unary_edge_length}
\end{figure*}

\subsection{Part Orientation Prediction}
\label{sec:supp_edge_direction}

{Fig.~\ref{fig:unary_rotation} illustrates the network architecture of our part orientation prediction module.
The input to the network is the concatenation of an image and a part mask.
A ResNet18 without pre-training followed by a convolution layer and an average-pooling layer is used to extract a part image feature.
The extracted image feature is forwarded to two fully-connected layers but with 4 different branches.
This illustrates the Mixture-of-Experts (MOE) method described in main paper Sec. 4.2 (Line 366-374), through which we hope to predict a distribution of rotations instead of a single output.
The final outputs of each branch are the quaternion and the probability to select this branch.}

The module is trained on the chair category for 20 epochs by the Adam optimizer.
The batch size is 64.
The initial learning rate is 0.001, and decays by 0.7 every 2 epochs.

\subsection{Part Size Prediction}
\label{sec:supp_edge_length}
For simplicity, we begin with the network architecture of single part size module which processes each part individually, and then describe how to extend it to the group-based size prediction.

Fig.~\ref{fig:unary_edge_length} shows the network architecture of the single part size prediction module. The goal is to predict the edge length of the bounding box along each axis from an image, a part mask, and principal axes. First, an image feature of a part is extracted with a 2D backbone, which is composed of a ResNet-18 without pre-training, a convolution layer, and an average-pooling layer. As our orientation module predicts principal axes of the part bounding box in an orderless fashion, our size module is designed to predict an edge length for each axis.
Inspired by PointNet~\cite{qi2017pointnet}, we consider each axis a `point', and estimate its attribute (edge length) in a point-wise segmentation fashion. Such a design is permutation-equivariant to the order of axes.
Concretely, we first concatenate the aggregated image feature and each axis as the input.
Then, we apply a shared multilayer perceptron (MLP) to extract the feature of each axis.
Following PointNet, a global feature is also extracted through the share MLP and concatenated with each local feature to enable the information communication between different axes.
Finally, a scalar edge length is estimated from the feature of each edge direction.

\begin{figure*}[t]
    \centering
    \includegraphics[width=\textwidth]{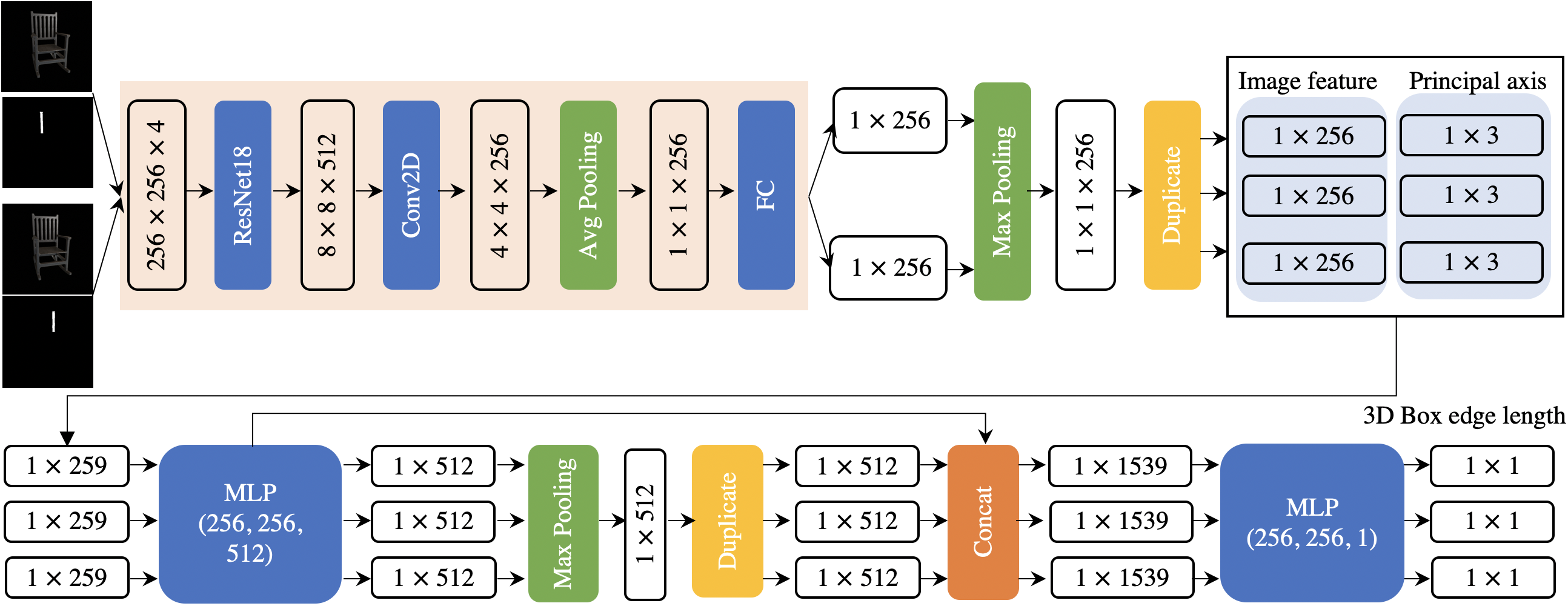}
    \caption{\titlecap{Network Architecture of the Group-based Part Size Prediction Module.}{The input to this module is a group of parts, which are considered to be translational symmetric. Single part image features are aggregated through max pooling, forming a grouped image feature. Then we predict edges length of the bounding box from this grouped image feature and part principal axes same as the single part size prediction.
       }}
    \label{fig:group_edge_length}
\end{figure*}

Fig.~\ref{fig:group_edge_length} illustrates the network architecture of the extension to the group-based prediction.
Given multiple parts that are considered translational symmetry, we use the 2D backbone of the single part size module to extract parts image features and then apply a max-pooling layer to aggregate all parts image features. Then we take the first part orientation as the uniform orientation for this translational symmetry part group. Again, we adopt the same architecture as the single part size module to predict a uniform size from the aggregated parts image feature and orientation.
During training, we leverage the ground truth translational symmetry relationship to generate the group.
During inference, it is based on the prediction of the part relationship module.

The module is trained on the chair category for 60 epochs by the Adam optimizer.
The batch size is 64.
The initial learning rate is 0.001 and decays by 0.9 every 2 epochs.

\subsection{Part Relative Position Prediction}
\label{sec:supp_relative_position}
\begin{figure*}[ht]
    \centering
    \includegraphics[width=\textwidth]{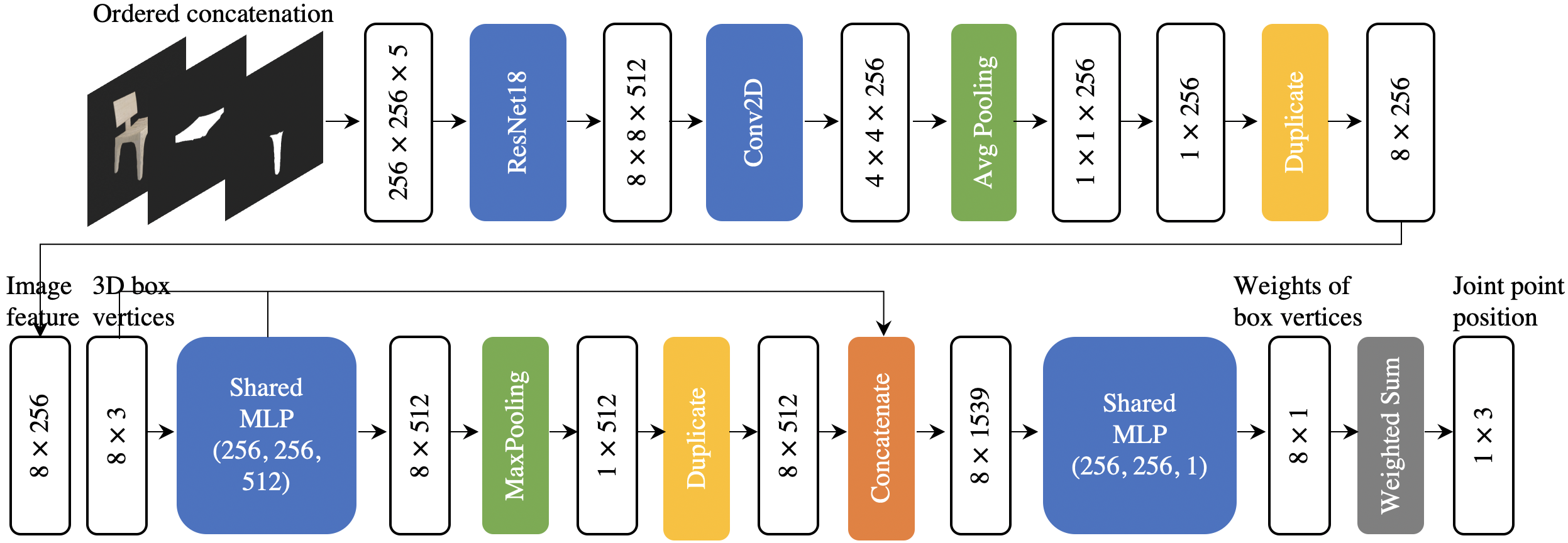}
    \caption{\titlecap{Network Architecture of the Contact-point-based Part Relative Position Prediction Module.}{This module estimates the contact point of both part given the image, part masks and part bounding box shape, \ie orientation and size. After getting both parts contact point, we infer the part relative position to the other part, as described in main paper Sec. 4.4 (Line 512-537). This contact-point module takes image and two adjacent part masks and part bounding box attributes, i.e.,  orientation and size. The contact point is represented as a weighted sum of all vertex position of the part bounding box. As the weight of each vertex is order-invariant to 3D box vertices, we take a similar architecture with PointNet. Each box vertex is considered as a point. We first concatenate the aggregated image feature and each box vertex as the input. Then, we apply a shared MLP to extract the feature of each edge direction. Following PointNet,a global feature is also extracted through the share MLP and concatenated with each local feature. Finally, each vertex weight is estimated from the feature of each box vertex.}}
    \label{fig:relative_position}
\end{figure*}

Fig.~\ref{fig:relative_position} shows the network architecture of our contact-point-based part relative position module. 
The network estimates the contact point of each part given the input image, part masks and the part bounding box shape, \ie the orientation and size. This contact point representation is then converted to a relative position, as explained in the main paper. As the process of estimating contact point is the same for both parts in a contacting pair, we illustrate the process for one of them as an example.

The 2D input is the concatenation of an image and masks of two adjacent parts. Note the masks are ordered where the first one corresponds to the part being considered and the second one corresponds to its contacting part.
A ResNet18 without pre-training followed by a convolution layer and an average-pooling layer is used to extract an image feature.

Then, the image feature is concatenated with each bounding box vertex of the part.
Similar to PointNet~\cite{qi2017pointnet} for semantic segmentation, the concatenated feature is processed by a shared MLP to obtain a weight for each bounding box vertex.
By summing all the vertex positions with their weights, we can induce the contact point of the part.
During training, the ground truth orientation and size are used to get the bounding box vertices. At testing stage, we take the predicted orientation and size from previous stages.

The module is trained on the chair category for 60 epochs by the Adam optimizer.
The batch size is 64.
The initial learning rate is 0.001, and decays by 0.9 every 2 epochs.

\section{Additional Ablation Studies}
\label{sec:ablation}
We provide additional ablation studies to validate the designs of some key network modules, complementing the two presented in the main paper.

\subsection{Part Orientation and Size:\\ Joint or Sequential Predictions?}
Our network design
disentangles the predictions of part 3D bounding box orientation and size.
Note that the order of estimated principal axes of part bounding box is tightly coupled with the order of edge length predictions along each axis.
As a result, in our final pipeline, we predict part orientation and size in a sequential manner.

One alternative way is to estimate them jointly as proposed in~\cite{mo2019structurenet}.
The joint prediction is supervised by the Chamfer Distance between the ground-truth part bounding box 
and the bounding box scaled by the predicted size and transformed by the predicted rotation.
Our experiments show that such joint prediction approach is more vulnerable to overfitting,
since it allows more degrees of freedom in the output space and exploits less explicit constraints compared to our approach.
%
Table~\ref{tab:ablate_rot_size} shows the quantitative comparison, where we clearly see the advantage of our sequential approach over the joint one.

\begin{table}[ht]
\centering
\begin{tabular}{lllll}
\toprule
 & Chair & Table & Cabinet & Bed \\
 \midrule
Joint      &  0.026 & 0.146 & 0.161 & 0.179  \\
Sequential (ours) &  \textbf{0.019} & \textbf{0.105} & \textbf{0.129} & \textbf{0.089}  \\
\bottomrule
\end{tabular}

\caption{\titlecap{Comparison of joint and sequential predictions of part orientation and size.}{We report the average Chamfer Distance between the ground truth and predicted part bounding boxes.}}
\label{tab:ablate_rot_size}
\end{table}

\subsection{Part Relative Position Prediction:\\ Center Offset or Contact Point?}
One alternative to our contact-point-based part relative position prediction method, which regresses the contact point of two parts, is to directly regress an offset vector from the center of one part to that of the other.
We conduct experiments to compare the performance of these two approaches.
Table~\ref{tab:ablate_joint} shows the quantitative comparison w.r.t the L1 distance between the ground truth and predicted offset vectors.
We argue that our contact-point-based part relative position prediction guarantees that two parts considered adjacent are contacted, which finally leads to superior performance.

\begin{table}[ht]
\centering
\begin{tabular}{lllll}
\toprule
 & Chair & Table & Cabinet & Bed \\ 
 \midrule
Center Offset &  0.027 & 0.042 & 0.051 &0.046 \\
Contact Point (ours) &  \textbf{0.023} & \textbf{0.038} & \textbf{0.045} & \textbf{0.042}  \\ 
\bottomrule
\end{tabular}
\caption{\titlecap{Comparison of Our Contact-point-based approach to an Center-offset-based Alternative.}{We report the average L1 distance between the ground truth and predicted offset vectors.}}
\label{tab:ablate_joint}
\end{table}

\end{document}